\documentclass[11pt]{article}

\usepackage[preprint]{acl}
\usepackage{times}
\usepackage{latexsym}
\usepackage{inconsolata}
\usepackage{stfloats}
\usepackage{fancyhdr}

\usepackage[utf8]{inputenc}
\usepackage[T1]{fontenc}
\usepackage{url}
\usepackage{xurl}
\usepackage{booktabs}
\usepackage{amsfonts}
\usepackage{nicefrac}
\usepackage{microtype}
\usepackage{xcolor}
\usepackage{listings}
\usepackage{tcolorbox}
\tcbuselibrary{listings,breakable,skins}
\usepackage{graphicx}
\usepackage{multirow}
\usepackage{array}
\usepackage{needspace}
\newcolumntype{L}[1]{>{\raggedright\arraybackslash}p{#1}}
\usepackage{amsmath}
\usepackage{amssymb}
\usepackage{flafter}
\usepackage{float}
\usepackage{placeins}

\newcommand{\benchmarkname}{\mbox{ICM-Bench}}
\newcommand{\cmark}{$\checkmark$}
\newcommand{\pmark}{$\triangle$}
\newcommand{\xmark}{--}

\title{ICM-Bench: Person-Level Identity Reasoning in Multimodal Agents with Long-Term Memory}

\author{%
  \textbf{Shidu Ren}\textsuperscript{1,2}\thanks{Work done during an internship at Memories.ai.} \quad
  \textbf{Yunze Liu}\textsuperscript{1,3}\thanks{Corresponding author.} \quad
  \textbf{Xing Liu}\textsuperscript{1,4} \\
  \textbf{Chi-Hao Wu}\textsuperscript{1} \quad
  \textbf{Enmin Zhou}\textsuperscript{1} \quad
  \textbf{Junxiao Shen}\textsuperscript{1,5} \\
  \normalfont\normalsize
  \textsuperscript{1}Memories.ai \quad
  \textsuperscript{2}University of Toronto \quad
  \textsuperscript{3}Tsinghua University \\
  \normalfont\normalsize\textsuperscript{4}Northeastern University \quad
  \textsuperscript{5}University of Bristol \\[2pt]
  \normalfont\normalsize\textbf{Code:} \url{https://github.com/Shidu-Ren/ICM-Bench}
}

\fancypagestyle{preprintfirst}{%
  \fancyhf{}%
  \fancyfoot[L]{\small Preprint}%
  \fancyfoot[C]{\thepage}%
}

\begin{document}
\maketitle
\thispagestyle{preprintfirst}

\begin{abstract}
Long-horizon multimodal agents should remember not only what happened but also who participated.
This capability depends on linking recurring faces, voices, names, person-associated objects, events, and social relations to consistent identities over time.
Existing long-video and multimodal-agent benchmarks measure broad memory question answering, but they do not isolate the ability to maintain recurring person identities and reason over their cross-time relations.
We introduce \benchmarkname{} (Identity-Centric Memory Benchmark), which, to the best of our knowledge, is the first benchmark specifically designed to evaluate identity-centric reasoning over long video memories in multimodal agents.
The benchmark contains 839 synthetic clips spanning 141 minutes and 1,217 open-ended questions about six recurring adults in a one-year life album.
A theme-configurable pipeline generates the video collection and associates each question with its target identities and traceable supporting evidence.
We compare direct caption-memory baselines, memory-augmented agents, and graph-retrieval systems.
Gemini 3.1 Pro achieves the highest overall accuracy of 74.0\%, yet its score falls to 60.3\% on questions that require long-term identity profiles.
The results show that current systems recover many event-level memories but remain less reliable when evidence must be accumulated around a stable person.
\end{abstract}

\begin{figure*}[t]
\centering
\includegraphics[width=\linewidth]{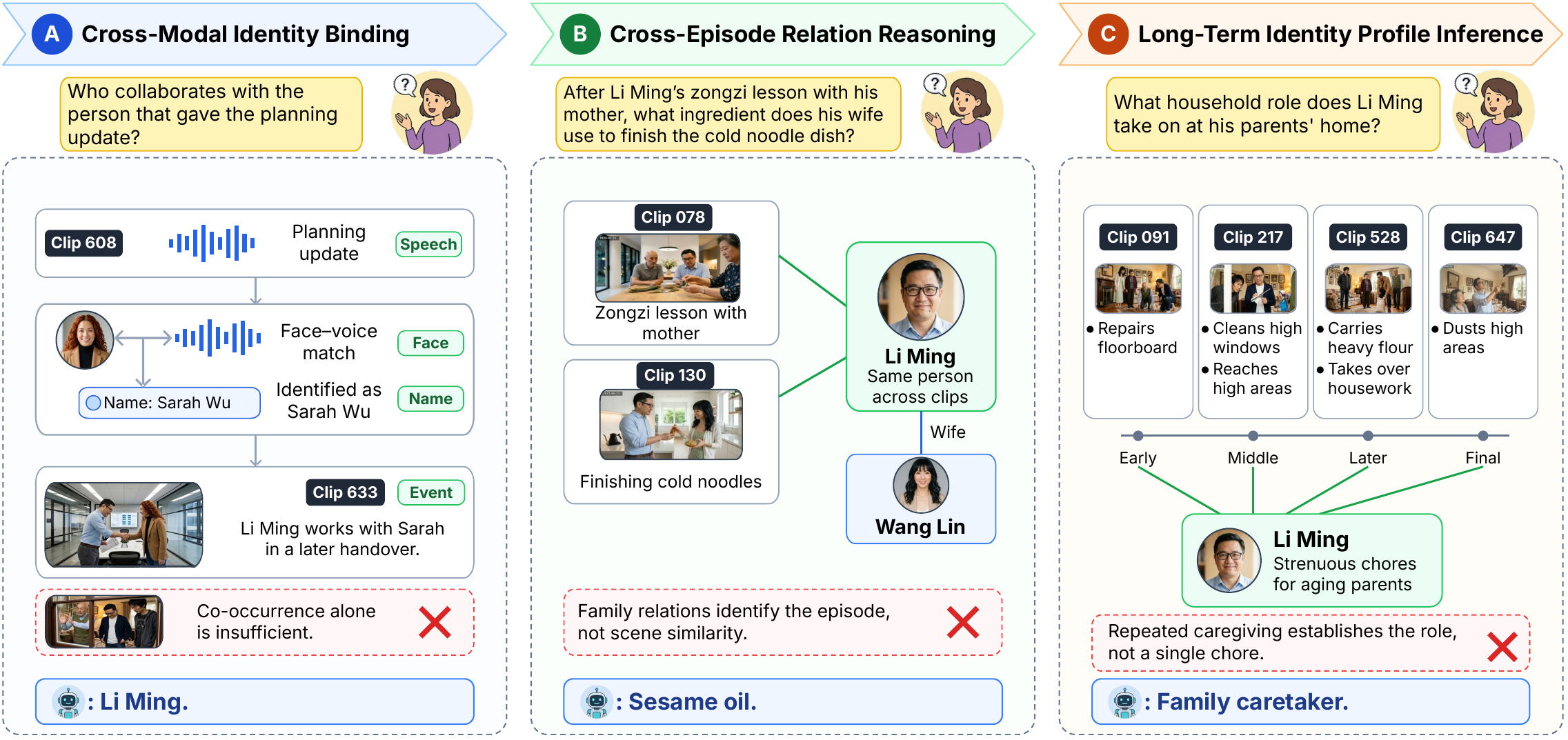}
\caption{\textbf{Three identity-centric memory challenges in \benchmarkname.}
Illustrative panels show how evidence is organized around recurring people across time, with (A) cross-modal binding of face, voice, name, and event cues, (B) relation reasoning across distant episodes, and (C) profile inference from repeated actions and relations.
The panels are illustrative rather than released benchmark items.
Red dashed boxes denote plausible but insufficient local cues or co-occurrences.}
\label{fig:overview}
\vspace{-8pt}
\end{figure*}

\section{Introduction}

Long-horizon multimodal agents need identity-centric relational memory: recognizing recurring people across time from faces, voices, names, objects, events, and social relations, then using these identities and relations to answer questions.
This capability underlies assistive agents, household robots, meeting assistants, and personal memory systems \citep{maharana2024locomo,kang-etal-2025-memory,long2026m3agent}, where ``remind me what she promised last week'' requires resolving the intended person from multimodal context.
Beyond face recognition and speaker clustering, identity-centric memory requires binding observations distributed across clips---a face, a voice, a name in a transcript, or a shared object---to stable person identities and reasoning over their associated events, roles, and relations.

Current systems represent long-term video memory as chronological captions and transcripts \citep{zhang2026silvr}, query-relevant records or segments \citep{lewis2020rag,kim2025salova}, or graphs connecting clips, events, passages, and entities \citep{shen2025vgent,gutierrez2025hipporag2}.
Multimodal memory agents further combine episodic, semantic, and visual stores or associate audio-visual observations through face, voice, and identity cues \citep{yeo2026worldmm,long2026m3agent}.
These designs reduce repeated processing and expose relevant evidence within limited context windows, while preserving different amounts of temporal, relational, and multimodal structure for reasoning over extended histories.

Recent benchmarks advance long-video evaluation through event understanding, temporal reasoning, evidence localization, and memory-based QA.
For example, Video-MME spans diverse domains and durations \citep{fu2025videomme}, LongVideoBench tests referring reasoning over interleaved video and subtitles \citep{wu2024longvideobench}, and M3-Bench evaluates agent memory, including person understanding and cross-modal reasoning \citep{long2026m3agent}.
However, aggregate QA scores do not reveal whether answers rely on stable identity reasoning or when identity mechanisms are necessary, helpful, or insufficient.
Direct evaluation of identity-centric memory therefore requires questions whose evidence depends on recurring people and their cross-time relations.

We introduce the Identity-Centric Memory Benchmark (\benchmarkname{}) to evaluate whether long-horizon multimodal agents can locate and aggregate evidence associated with recurring people and their cross-time relations.
To the best of our knowledge, \benchmarkname{} is the first benchmark designed specifically to evaluate this capability over long video memories.
Figure~\ref{fig:overview} illustrates its three core challenges: cross-modal identity binding, cross-episode relation reasoning, and long-term identity profile inference.
A controllable generation pipeline creates synthetic memories with recurring adults, persistent appearance references, voice-consistent dialogue, and temporally distributed identity and relation cues.
Open-ended QA and structured annotations specify categories, target identities, and supporting clips and shots, while the evaluation protocol controls each question's accessible timeline.

Using \benchmarkname{}, we evaluate closed-source and open-source direct caption-memory baselines, memory-augmented and graph-retrieval systems, and Automatic Speech Recognition (ASR)-only controls that answer from transcripts without visual captions or frames.
The strongest direct baseline, Gemini 3.1 Pro \citep{google2026gemini31pro}, reaches 74.0\% overall accuracy but only 60.3\% on long-term identity-profile questions, indicating that strong models can recover evidence while profile inference remains difficult.
Several memory-augmented and retrieval settings improve over their corresponding open-source direct baselines, whereas ASR-only controls fall well below caption-based memory, indicating that many questions require visual evidence distributed across the timeline.

Our contributions are summarized as follows:
\vspace{-4pt}
\begin{enumerate}
    \item \textbf{Benchmark.} We introduce \benchmarkname{}, comprising long-form synthetic audio-visual memory and 1,217 traceable, temporally controlled QA pairs for identity-centric evidence retrieval and cross-time relation reasoning.\vspace{-4pt}
    \item \textbf{Theme-configurable generation pipeline.} We develop a reusable pipeline that constructs identity-consistent audio-visual memories and evidence-grounded open-ended QA from configurable themes.\vspace{-4pt}
    \item \textbf{Diagnostic evaluation.} We evaluate direct model baselines, representative memory-augmented frameworks, and ASR-only transcript controls under a unified open-ended QA judge. The analysis separates event-level memory from accumulated identity-profile inference and shows that transcript-only memory is insufficient.
\end{enumerate}

\section{Related work}
\label{sec:related-work}

\begin{table*}[t]
\centering

\small
\setlength{\tabcolsep}{3pt}
\begin{tabular*}{\linewidth}{@{\extracolsep{\fill}}lccccc@{}}
\toprule
\textbf{Configuration} & \textbf{Face} & \textbf{Video} & \textbf{Audio} & \shortstack{\textbf{Face--voice}\\\textbf{link}} & \shortstack{\textbf{Acc.}\\\textbf{(\%)}} \\
\midrule
Full pipeline & \cmark & \cmark & \cmark & \cmark & 52.5 \\
w/o face processing & -- & \cmark & \cmark & N/A & 55.3 \\
Audio only & -- & -- & \cmark & N/A & 52.2 \\
w/o face--voice linking & \cmark & \cmark & \cmark & -- & 53.0 \\
\bottomrule
\end{tabular*}
\caption{\textbf{M3-Agent identity-processing diagnostic on M3-Bench Web.}
Results are evaluated on 575 questions from 100 randomly sampled videos.}
\label{tab:pilot-m3}
\vspace{-8pt}
\end{table*}

\textbf{Long-video and multimodal-memory benchmarks.}
Long-video QA benchmarks such as EgoSchema, Video-MME, LongVideoBench, and InfiniBench evaluate temporal understanding and multimodal reasoning over extended videos \citep{mangalam2023egoschema,fu2025videomme,wu2024longvideobench,ataallah-etal-2025-infinibench}.
Recent memory benchmarks cover long-term conversations, personalized references, multi-session multimodal evidence, entity and event memory, and evolving world states \citep{maharana2024locomo,mei2026atmbench,ren2026memlens,guo2026memeye,liu2026worldmemarena}.
EgoLife includes identity-aware life QA, habit understanding, and relation reasoning grounded in week-long recordings \citep{yang2025egolife}, while EgoMemReason evaluates entity, event, and behavior memory across days, including changing states and recurring behavioral patterns \citep{wang2026egomemreason}.
Controlled diagnostic benchmarks such as CLEVR and GQA provide a precedent for targeting specific reasoning capabilities through structured construction \citep{johnson2017clevr,hudson2019gqa}.
\benchmarkname{} centers its questions on evidence associated with recurring people and their cross-time relations.
Its controlled synthetic audio-visual timeline supports identity recall, cross-episode retrieval, and long-term profile inference, with question-specific access limits and traceable supporting clips and shots.

\textbf{Long-horizon multimodal agents and memory organization.}
Long-term multimodal systems convert accumulating audio-visual experience into persistent representations that can be searched and updated for later reasoning \citep{long2026m3agent,yeo2026worldmm}.
M3-Agent organizes episodic and semantic memories around entities and links face and speaker nodes inferred to refer to the same person \citep{long2026m3agent}, while WorldMM retrieves from complementary episodic, semantic, and visual memories at multiple temporal scales \citep{yeo2026worldmm}.
Retrieval-augmented generation provides access to external memory \citep{lewis2020rag}.
Graph-based approaches organize video clips and events in Vgent, or passages and entities in HippoRAG2, to support retrieval and reasoning \citep{shen2025vgent,gutierrez2025hipporag2}.

\textbf{Person identity grounding across modalities.}
Face recognition and person re-identification associate visual observations with identity, while speaker recognition and diarization perform the corresponding task for speech \citep{schroff2015facenet,deng2019arcface,zheng2015scalable,luo2019bagtricks,snyder2018xvectors,desplanques2020ecapa,bredin2020pyannote}.
Cross-modal methods further match voices to faces or associate speech with visible speakers \citep{nagrani2018learnablepins,roth2020avaactive}.
AMusE evaluates speaker association, re-identification across non-contiguous clips, and cross-scene narrative linking \citep{chowdhury2025amuse}.
M3-SLU tests speaker-attributed QA and utterance matching in multi-turn conversations \citep{kwon2025m3slu}.
\benchmarkname{} evaluates person-centered QA over a long audio-visual timeline, combining event retrieval, relation reasoning, and profile inference from accumulated identity-linked evidence.
Profile questions target roles, habits, preferences, and relationship patterns inferred from repeated observations.
\section{\benchmarkname{}}
\label{sec:icm-bench}

\subsection{Motivating identity-processing diagnostic}
\label{sec:pilot-study}

To test whether aggregate performance on a broad multimodal-memory benchmark reflects explicit person-identity processing, we conduct a controlled diagnostic with M3-Agent on the Web subset of M3-Bench. M3-Agent is useful for this purpose because it builds video and audio memories with separate face and speaker representations and uses equivalence links to associate evidence that may refer to the same recurring person \citep{long2026m3agent}. If broad benchmark accuracy depends on these identity mechanisms, removing dedicated face processing or face--voice equivalence linking should reduce QA performance. We randomly sample 100 videos and include all 575 questions associated with them, then compare four variants: the full \textit{Face+Video+Audio} pipeline, \textit{Video+Audio} without face processing, \textit{Audio only} without either face processing or video input, and the full-input pipeline with face--voice equivalence linking disabled.

As shown in Table~\ref{tab:pilot-m3}, aggregate QA accuracy is largely preserved and can even improve when explicit identity components are removed. Removing face processing raises accuracy from 52.5\% to 55.3\%, while disabling face--voice equivalence linking yields 53.0\%, which also exceeds the full pipeline. The \textit{Audio only} setting removes the video stream yet retains 52.2\% accuracy. Broad aggregate accuracy can therefore remain largely unchanged without dedicated visual identity processing because many questions are answerable from transcripts, local events, or co-occurrence cues. \benchmarkname{} instead centers its questions on recurring people and cross-time relations, making performance sensitive to how systems use person-linked evidence.

\subsection{Task definition and benchmark annotations}

We formulate \benchmarkname{} as open-ended QA over chronological audio-visual memory in which the same people recur across episodes. Systems must locate and aggregate evidence associated with a queried person over time. The benchmark sequence comprises chronological clips $\mathcal{V}=\{c_1,\ldots,c_T\}$ and a recurring cast $\mathcal{I}$. Identity Recall and Cross-Episode Identity Retrieval answer a question $x$ using evidence from the prefix $\mathcal{V}_{\leq\tau}$, whereas Long-Term Identity Profile Inference uses the full timeline to infer a person's role, habits, preferences, or relationships from observations accumulated over time. Table~\ref{tab:benchmark-positioning} contrasts this evaluation target with related benchmarks.

\begin{table*}[t]
\centering

\small
\setlength{\tabcolsep}{3.5pt}
\begin{tabular*}{\textwidth}{@{\extracolsep{\fill}}lcccccc}
\toprule
\textbf{Benchmark / line} & \textbf{Long ctx.} & \shortstack{\textbf{Audio-}\\\textbf{visual}} & \shortstack{\textbf{Agent}\\\textbf{mem.}} & \shortstack{\textbf{Visual identity}\\\textbf{continuity}} & \shortstack{\textbf{Identity}\\\textbf{evidence}} & \shortstack{\textbf{Traceable}\\\textbf{anchors}} \\
\midrule
EgoSchema & \cmark & \xmark & \xmark & \pmark & \xmark & \xmark \\
Video-MME & \cmark & \cmark & \xmark & \pmark & \pmark & \pmark \\
LongVideoBench & \cmark & \pmark & \xmark & \pmark & \pmark & \pmark \\
InfiniBench & \cmark & \cmark & \xmark & \cmark & \pmark & \pmark \\
EgoLife / EgoLifeQA & \cmark & \cmark & \cmark & \pmark & \pmark & \cmark \\
EgoMemReason & \cmark & \pmark & \cmark & \pmark & \pmark & \pmark \\
AMUSE & \pmark & \cmark & \pmark & \cmark & \cmark & \pmark \\
M3-SLU & \pmark & \xmark & \xmark & \xmark & \cmark & \pmark \\
M3-Bench & \cmark & \cmark & \cmark & \pmark & \pmark & \pmark \\
LoCoMo & \cmark & \xmark & \cmark & \xmark & \pmark & \cmark \\
\benchmarkname{} & \cmark & \cmark & \cmark & \cmark & \cmark & \cmark \\
\bottomrule
\end{tabular*}
\caption{\textbf{Positioning of \benchmarkname{} against related evaluation lines.}
The first three columns cover context, modality, and agent memory. Visual identity continuity denotes support for visually matching the same person across separated scenes or clips through face and appearance cues. The final two columns indicate identity-linked evidence requirements and question-level supporting anchors. Symbols denote central coverage (\cmark), partial or incidental coverage (\pmark), and no primary coverage (\xmark). Face anonymization limits visual identity cues in EgoLife and EgoMemReason.}
\label{tab:benchmark-positioning}
\vspace{-8pt}
\end{table*}

Each item is ${q_i=(x_i,a_i,y_i,\mathcal{E}_i,\mathcal{I}_i,\tau_i)}$, where $x_i$ and $a_i$ are the question and answer, $y_i$ is its category, $\mathcal{E}_i$ indexes supporting clips or shots, $\mathcal{I}_i$ identifies the people involved, and $\tau_i$ is the latest clip permitted in the answer context, with $\tau_i=T$ for Profile questions. Profile questions test whether an agent can characterize a recurring person from observed behavior for personalized reminders, recommendations, or assistance. Most questions name the queried person, while supporting clips associate that person with face, voice, name, object, or relation cues. The system returns a free-form answer $\hat{a}$, and the semantic-equivalence judge in Section~\ref{sec:experiments} accepts answers that convey the reference meaning rather than requiring identical wording. Reference answers, category labels, target-identity annotations, and supporting-evidence annotations are withheld from systems; the cutoff controls the answer context. \benchmarkname{} does not prescribe a memory architecture: systems may store captions, entity or face--voice records, or graph structures and select answer evidence within the cutoff, while only the final answer contributes to the primary metric.

\subsection{Theme-configurable benchmark construction}
\label{sec:generation-pipeline}

\begin{figure*}[t]
\centering
\includegraphics[width=\linewidth]{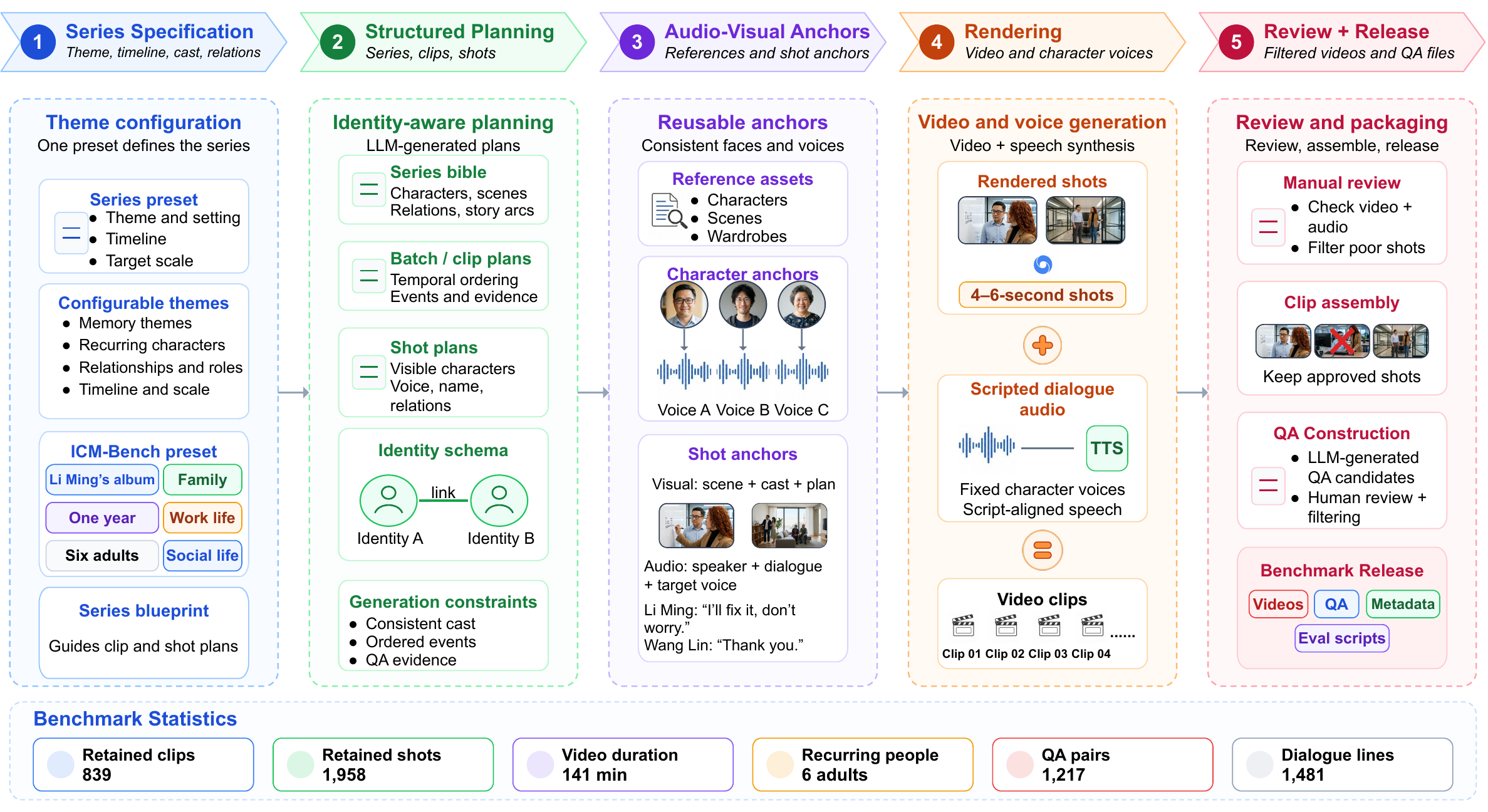}
\caption{\textbf{Theme-configurable generation pipeline for \benchmarkname.}
A series-level theme preset is expanded into planning metadata, reusable audio-visual anchors, rendered video and dialogue, quality control, and benchmark files.
Because each stage retains its planning and review records, the final questions can be traced to the intended identities, relations, and supporting shots.}
\label{fig:pipeline}
\vspace{-8pt}
\end{figure*}

To control person-level evidence across a long video series, \benchmarkname{} separates story planning, visual anchoring, rendering, voice redubbing, review, and QA construction. Figure~\ref{fig:pipeline} summarizes how a theme preset defines the timeline, recurring cast, dialogue, rendering settings, and QA distribution. Retained planning and review records link final questions to the intended identities, relations, and supporting shots, providing curation provenance without being exposed to evaluated systems.

The planning stage determines where recurring characters and identity evidence will appear across the series before any video is rendered. A planning model produces a series specification defining the cast, settings, and relationships, along with clip outlines and wardrobe plans that place characters, relations, and scenes in temporal order. A deterministic stage then allocates shot counts, durations, visible cast, dialogue, and planned identity and event cues, after which a refinement model produces final shot plans locating face, voice, name, object, and relation cues. Together, these plans distribute identity anchors across the timeline and separate them from answer-bearing events.

The rendering stage turns reviewed visual references and shot plans into identity-consistent video clips. Visual anchors are first checked for identity consistency, scene match, composition, and visible artifacts, and only accepted anchors proceed to rendering. Reviewed character, scene, and shot references are then reused across batches to generate primarily four- or six-second, 16:9, 720p image-to-video shots, while character-specific speech synthesis maintains recurring voices. After audio redubbing and mixing, accepted shots are assembled chronologically as date-stamped clips that retain visible dates but omit subtitles and construction labels that could reveal answers. Separating theme configuration from the benchmark schema allows the same process to instantiate a different life-album setting without changing the evaluation protocol.

\begin{table*}[!t]
\centering

\small
\setlength{\tabcolsep}{3pt}
\begin{tabular}{L{0.22\textwidth}L{0.24\textwidth}L{0.27\textwidth}L{0.21\textwidth}}
\toprule
Category & Target operation & Evidence design & Diagnostic focus \\
\midrule
Identity Recall &
Recover a local fact attached to the correct recurring person. &
Exactly one evidence clip with an explicit person, object, action, or relation cue. &
Whether the event is attached to the correct recurring person. \\
Cross-Episode Identity Retrieval &
Link separated clips through a recurring identity or relation. &
At least two temporally separated evidence clips, with face, voice, name, or relation anchors distributed over time. &
Whether separated clips are linked through the same identity or relation. \\
Long-Term Identity Profile Inference &
Infer a stable role, habit, preference, or relationship pattern. &
Repeated evidence across the series, often involving family, work, or social roles. &
Whether repeated evidence is aggregated into a stable person-level profile. \\
\bottomrule
\end{tabular}
\caption{\textbf{\benchmarkname{} QA taxonomy.}
The three families test local recall, cross-episode retrieval, and long-term profile inference. The final column identifies the main reasoning ability each question family is designed to probe.}
\label{tab:qa-taxonomy-main}
\vspace{-8pt}
\end{table*}

We instantiate the pipeline as \textit{Li Ming's Life Album}, a one-year synthetic album from May 1, 2025 to May 1, 2026, with six recurring adults and 44 scene groups. Some public and social scenes also contain unnamed, non-recurring adult background figures, which are not treated as benchmark identities. Human reviewers compare each shot with its intended cast, action, dialogue, and visual anchors and remove shots with identity drift, action mismatch, severe artifacts, audio failure, or unintended shortcuts. The final 141.0-minute sequence contains 839 clips: a short calibration clip that links the recurring adults' faces and voices, followed by 838 date-stamped memory clips spanning 140.6 minutes. Representative questions appear in Appendix~\ref{app:qa-examples}.

To test whether the construction process transfers to another theme, we apply the same schema to \textit{Backstage in S\~ao Paulo}, a theater-rehearsal series with a different cast and setting. The resulting series contains six recurring adults, 14 scene groups, 30 clips, and 84 planned shots. It is excluded from benchmark statistics and experiments and is used only to verify reuse of the schema under a different theme. Representative construction artifacts from this pilot are shown in Appendix~\ref{app:second-theme-example}.

Gemini 3.1 Pro supports planning, anchor checking, and QA candidate generation, while Gemini 3 Flash refines shot plans and audits Profile candidates. Gemini 3.1 Flash Image provides visual references, Veo 3.1 Fast Generate renders shots, and Gemini 3.1 Flash TTS supplies recurring voices \citep{google2026gemini31pro,google2026gemini3flash,google2026gemini31flashimage,google2025veo31,google2026gemini31flashtts}.

\vspace{-6pt}
\subsection{QA construction and quality control}
\vspace{-4pt}

QA construction proceeds through candidate generation, automated filtering, and manual review. An independent stage creates open-ended questions from retained shot plans, visual events, dialogue, timeline facts, and character relations rather than captions produced by evaluated systems. Each candidate records its category, target identities, reference answer, supporting evidence, and cutoff. Structured checks reject malformed, temporally invalid, duplicate, and category-inconsistent items, while a separate audit removes semantically redundant questions and checks whether Profile candidates require accumulated person-level inference. Two authors review every retained question against its permitted audio-visual memory, verifying answerability, the reference answer, annotated evidence, and cutoff. Only items passing every check are retained; Appendix~\ref{app:qa-quality-control} details the review and disagreement-resolution procedure. The final open-ended format requires systems to produce the relevant person, relation, object, or profile rather than select among distractor options. The resulting set contains 1,217 questions: 400 Identity Recall, 500 Cross-Episode Identity Retrieval, and 317 Long-Term Identity Profile Inference, abbreviated as Recall, Retrieval, and Profile in Table~\ref{tab:qa-taxonomy-main}.

\section{Experiments}
\label{sec:experiments}

\begin{table*}[t]
\centering

\small
\setlength{\tabcolsep}{3.5pt}
\begin{tabular*}{\textwidth}{@{\extracolsep{\fill}}lllcccc}
\toprule
Systems & Method / Variant & Model & Overall & Recall & Retrieval & Profile \\
\midrule
\multirow{1}{*}{Human}
& -- & -- & 93.8 & 96.8 & 94.2 & 89.6 \\
\midrule
\multirow{4}{*}{Direct Models}
& Caption-memory & Gemini 3.5 Flash & 73.4 & \textbf{83.8} & 79.4 & 50.8 \\
& Caption-memory & Gemini 3.1 Pro & \textbf{74.0} & 76.8 & \textbf{80.6} & \textbf{60.3} \\
& Caption-memory & Qwen3.5-9B & 47.9 & 53.5 & 45.8 & 44.2 \\
& Caption-memory & Qwen3-VL-8B-Instruct & 39.4 & 44.3 & 33.2 & 43.2 \\
\midrule
\multirow{6}{*}{Memory Agents}
& Vgent & Qwen3.5-9B & 52.3 & 66.0 & 54.6 & 31.2 \\
& Vgent & Qwen3-VL-8B-Instruct & 40.3 & 47.8 & 39.2 & 32.8 \\
& M3-Agent & Gemini 3.5 Flash & 58.7 & 67.0 & 60.6 & 45.1 \\
& M3-Agent (Original) & Qwen2.5-Omni-7B (SFT) & 45.4 & 55.5 & 50.8 & 24.3 \\
& M3-Agent no equiv & Qwen2.5-Omni-7B (SFT) & 42.9 & 49.3 & 47.8 & 27.1 \\
& M3-Agent no face & Qwen2.5-Omni-7B (SFT) & 41.0 & 49.3 & 45.0 & 24.3 \\
\midrule
\multirow{3}{*}{RAG Systems}
& HippoRAG2 & Gemini 3.5 Flash & 63.4 & 72.5 & 70.4 & 40.7 \\
& HippoRAG2 & Qwen3.5-9B & 49.6 & 61.3 & 54.4 & 27.4 \\
& HippoRAG2 & Qwen3-VL-8B-Instruct & 53.3 & 57.0 & 57.6 & 42.0 \\
\midrule
\multirow{2}{*}{ASR-only}
& Direct memory & Gemini 3.5 Flash & 38.7 & 37.3 & 44.2 & 31.9 \\
& HippoRAG2 & Qwen3.5-9B & 29.9 & 34.8 & 34.0 & 17.4 \\
\bottomrule
\end{tabular*}
\caption{\textbf{Main \benchmarkname{} results on 1,217 open-ended questions.}
Values are answer accuracy in percent under the shared open-ended judging protocol.
Overall averages all questions, while Recall, Retrieval, and Profile correspond to the three QA families defined in Table~\ref{tab:qa-taxonomy-main}, with rows grouped by system family.
The Human row reports mean answer accuracy and is excluded when bolding automated scores.}
\label{tab:main-results}
\vspace{-10pt}
\end{table*}

\subsection{Experimental setup}

\textbf{Dataset.}
The evaluation uses all 1,217 questions in \benchmarkname{}, including 400 Identity Recall, 500 Cross-Episode Identity Retrieval, and 317 Long-Term Identity Profile Inference questions. The underlying sequence contains 839 clips, beginning with the calibration clip described in Section~\ref{sec:generation-pipeline} and followed by 838 date-stamped memory clips spanning 140.6 minutes. For Recall and Retrieval questions, the evidence supplied to the answerer is restricted to clips at or before the question-specific cutoff. Profile questions use the full timeline to infer a person's role, habits, preferences, or relationships from observations accumulated over time.

\textbf{Baselines.}
We compare direct caption-memory models, long-term memory agents, graph-retrieval systems, and transcript-only controls. The direct group includes Gemini 3.5 Flash, Gemini 3.1 Pro, Qwen3.5-9B, and Qwen3-VL-8B-Instruct \citep{google2026gemini35flash,google2026gemini31pro,qwen3.5,qwen2026qwen3vl8b}. The memory-agent group includes Vgent and M3-Agent, together with an M3-Agent variant that uses Gemini 3.5 Flash and two ablations of the released Qwen2.5-Omni-7B setting \citep{long2026m3agent,shen2025vgent,xu2025qwen25omni}. HippoRAG2 is evaluated with Gemini 3.5 Flash and both Qwen answerers \citep{gutierrez2025hipporag2}. We also report human answer accuracy over all 1,217 questions as a post-hoc answerability reference, not a formal human-performance ceiling.

\textbf{Metrics.}
We report answer accuracy over the full benchmark and separately for Recall, Retrieval, and Profile questions. Since answers are free-form, a shared Gemini 3 Flash judge determines whether the reference answer is entailed by the model response \citep{google2026gemini3flash}. This semantic-equivalence protocol follows prior work on open-ended model evaluation \citep{zheng2023judging,liu2023geval,long2026m3agent,mei2026atmbench}, with the complete prompt given in Appendix~\ref{app:prompt-templates}.

\textbf{Implementation details.}
All systems receive the same question and answer-format instructions with only the wrappers required by each framework. Gemini direct baselines caption complete clips, while Qwen direct baselines caption 12 uniformly sampled frames together with ASR text. Memory-agent and HippoRAG2 runs use their native memory construction and retrieval procedures, whereas the transcript-only controls remove frames, visual captions, face cues, and image-derived event descriptions. Every video-based setting receives the calibration clip described in Section~\ref{sec:generation-pipeline}, and the primary judge runs at temperature $10^{-6}$.

\subsection{Main results}

\begin{figure}[t]
\centering
\includegraphics[width=\linewidth]{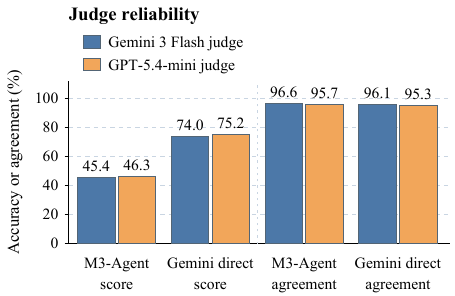}
\caption{\textbf{Judge reliability.}
Overall QA scores assigned by Gemini 3 Flash and GPT-5.4-mini. Agreement bars average each judge's agreement with two independently produced human label sets.}
\label{fig:response-audit}
\vspace{-12pt}
\end{figure}

\textbf{Human answers remain substantially more accurate than automated systems.} Table~\ref{tab:main-results} presents the results under the shared open-ended QA protocol. Human accuracy is 93.8\% overall, while the best automated system, Gemini 3.1 Pro with direct caption memory, reaches 74.0\% overall and 60.3\% on Profile questions. Memory-agent and graph-retrieval frameworks improve several open-source settings but do not surpass direct Gemini caption memory, while ASR-only controls perform much worse because transcripts omit visual evidence required by many identity-centric questions. Separately, we rescore two complete answer runs with GPT-5.4-mini \citep{openai2026gpt54mini}. On the final 1,217 questions, the two judges produce similar overall scores for both audited runs. Both judges show 95.3--96.6\% mean agreement with two independently verified human label sets, as shown in Figure~\ref{fig:response-audit}.

\textbf{Memory organization can improve event retrieval while weakening Profile inference.} The pattern varies by answerer: direct captions work best with Gemini, whereas Vgent and HippoRAG2 improve Qwen baselines overall but lower their Profile scores. With Gemini 3.5 Flash, direct caption memory reaches 73.4\%, compared with 63.4\% for HippoRAG2 and 58.7\% for M3-Agent. Thus, when a strong long-context model can use the available captions, compressing or restructuring them does not improve answer accuracy in this evaluation. The open-source results show the category tradeoff: Profile accuracy falls from 44.2\% to 31.2\% with Vgent and 27.4\% with HippoRAG2 for Qwen3.5-9B, and from 43.2\% to 32.8\% and 42.0\% for Qwen3-VL-8B-Instruct, despite overall gains. Persistent memory frameworks may still reduce repeated processing or provide reusable state for future updates, but improvements in event-oriented Recall and Retrieval do not necessarily extend to aggregating evidence into stable person-level profiles.

\textbf{Identity-related components affect performance on \benchmarkname{}.} The original M3-Agent achieves 45.4\% overall accuracy, compared with 42.9\% without equivalence linking and 41.0\% without face processing. Both ablated variants score below the original system, showing that dedicated face processing and equivalence linking each contribute to overall QA accuracy in this setting. The full system nevertheless makes identity-linking errors, as illustrated by the retrieved-memory example in Appendix~\ref{app:m3-link-diagnostic}. In this case, observations of different people are incorrectly associated with the same identity in the system's memory. Together, the ablations and this example illustrate how \benchmarkname{} reveals the contribution of identity-related components to overall QA accuracy and exposes incorrect identity associations that remain in the full system.

\textbf{ASR-only memory is insufficient.} ASR-only controls give systems only transcript memory for answering, removing visual captions, frames, face cues, and image-derived event descriptions. Replacing caption memory with ASR-only memory drops Gemini 3.5 Flash from 73.4\% to 38.7\% overall, while HippoRAG2 with Qwen3.5-9B reaches 29.9\% in the ASR-only setting. This comparison measures the contribution of non-transcript visual information, and the gaps show that speech transcripts alone are insufficient for many questions.

\textbf{Profile remains difficult.} The best automated Profile score is 60.3\%, far below the human score of 89.6\%, and most open-source or framework settings remain below 46\%. Unlike Recall and Retrieval, Profile questions require aggregating repeated identity, role, and relationship evidence across the album before producing a stable person-level answer. This gap is the main diagnostic value of \benchmarkname{} because high event-level accuracy does not imply stable person-level memory.

\subsection{Qualitative Failure Analysis}

Qualitative inspection of Profile errors reveals three illustrative failure modes that help explain the gap between event-level recall and person-level inference. In a relationship-specific profile miss, M3-Agent resolves Li Ming's identity and retrieves photography-related memories, but answers ``candid street photography'' instead of identifying Wang Lin's candid, unposed moments as the recurring subject. In a retrieval-distractor case, Vgent returns memories that are topically related to travel but answers ``travel receipts or tickets'' instead of identifying the postcards repeatedly associated with the same couple. None of its five retrieved memories mentions postcards, while the fifth explicitly mentions tickets and receipts that appear in the answer. The third pattern is incomplete cross-clip evidence integration: Vgent names a watering can and shears but omits the auto-watering bulbs explicitly mentioned in its retrieved gardening memories. In these examples, the systems can describe individual scenes but fail to organize them around a stable person, relationship, or recurring object. The cases complement the aggregate Profile scores in Table~\ref{tab:main-results} by showing that failures can occur during relation interpretation, evidence selection, or cross-episode aggregation rather than at a single common stage. Appendix~\ref{app:failure-analysis} gives the question, evidence pattern, reference answer, representative output, and diagnosis for one example of each failure mode.

\section{Conclusion}

We introduced \benchmarkname{}, an identity-centric benchmark that combines reviewed synthetic life-album clips, traceable person-level annotations, and 1,217 open-ended questions to evaluate recurring identities and cross-time relations in long-horizon multimodal memory. The benchmark covers identity recall, cross-episode retrieval, and long-term profile inference under a shared evaluation protocol. Across direct models, memory agents, and graph-retrieval systems, the strongest automated system reaches 74.0\% overall but only 60.3\% on Profile questions, while transcript-only controls perform substantially worse. These results show that current systems can recover many individual events but still struggle to integrate audio-visual evidence across episodes about the same individuals. We hope \benchmarkname{} will support the development of multimodal agents with reliable and personalized long-term memory.

\label{arr:main-end}
\section*{Limitations}
\label{arr:limitations-start}
 \benchmarkname{} uses reviewed synthetic audio-visual content, which may contain visual artifacts, imperfect motion, cross-shot inconsistencies, or unnatural dialogue and omits real-video challenges such as occlusion, camera motion, changing illumination, background noise, and overlapping speech. Results should therefore be viewed as a controlled diagnostic rather than evidence of transfer to unconstrained personal recordings. Future work should extend the benchmark to consented real-world video and broader social settings while preserving traceable identity evidence and privacy safeguards.

\section*{Ethical considerations}
\label{arr:ethics-start}
The main \benchmarkname{} album depicts synthetic adults rather than recorded participants.
Applying identity-centric memory to real recordings raises risks of surveillance and sensitive profiling through faces, voices, relationships, and habits.
Such applications should obtain informed consent, minimize stored personal data, and provide access controls and deletion mechanisms.
Synthetic generation may also reproduce stereotypes, and our limited cast does not establish demographic or cultural coverage.

Human curation and evaluation were performed by authors who consented to participate and received no additional compensation.
The answerability and judge-reliability evaluators were separate from the benchmark curators, as described in Appendix~\ref{app:human-reference}.

Our original code is released under MIT, with third-party components retaining their upstream licenses.
\benchmarkname{} and M3-Bench are distributed under CC BY-NC-SA 4.0.
For the M3-Bench Web diagnostic, source-video rights and platform terms remain applicable in addition to the dataset license.

\label{arr:references-start}
\bibliography{references}

\appendix
\label{arr:appendix-start}

\section*{Appendices}
\subsection*{Contents}
\begingroup
\makeatletter
\newcommand{\appendixtocentry}[1]{%
  \@dottedtocline{1}{0em}{1.5em}%
    {\hyperref[#1]{\numberline{\ref*{#1}}\nameref*{#1}}}%
    {\hyperref[#1]{\pageref*{#1}}}%
}
\appendixtocentry{app:implementation-details}
\appendixtocentry{app:construction-examples}
\appendixtocentry{app:qa-examples}
\appendixtocentry{app:qa-quality-control}
\appendixtocentry{app:human-reference}
\appendixtocentry{app:judge-reliability}
\appendixtocentry{app:prompt-templates}
\appendixtocentry{app:m3-link-diagnostic}
\appendixtocentry{app:failure-analysis}
\makeatother
\endgroup

\lstset{
  basicstyle=\ttfamily\small,
  breaklines=true,
  breakatwhitespace=false,
  columns=fullflexible,
  keepspaces=true,
  frame=none,
  xleftmargin=0pt,
  xrightmargin=0pt,
  aboveskip=0.4em,
  belowskip=0.6em
}

\tcbset{
  icmbox/.style={
    enhanced jigsaw,
    breakable,
    colback=white,
    colframe=black!45,
    boxrule=0.45pt,
    arc=1pt,
    left=7pt,
    right=7pt,
    top=5pt,
    bottom=5pt,
    fonttitle=\small\bfseries,
    coltitle=black,
    colbacktitle=black!7
  }
}

\newtcblisting{promptbox}[1][]{
  icmbox,
  breakable,
  ignore nobreak=true,
  before skip=0pt,
  top=2pt,
  left=5pt,
  right=5pt,
  title={#1},
  title after break={#1 (continued)},
  listing only,
  listing options={
    basicstyle=\ttfamily\small,
    breaklines=true,
    breakatwhitespace=false,
    columns=fullflexible,
    keepspaces=true,
    breakautoindent=false,
    breakindent=0pt,
    emptylines=0,
    aboveskip=0pt,
    belowskip=0pt
  }
}

\section{Implementation details}
\label{app:implementation-details}

\textbf{Shared evaluation interface.}
All methods are evaluated on the same 1,217 questions and receive the same concise-answer instruction. Evaluator-side fields, including the reference answer, target identities, and evidence annotations, are never exposed to the answering system. Every video-based method may use the introductory calibration clip, which supplies a common face--voice reference without revealing names. For Recall and Retrieval questions, the answer context contains only evidence from clips at or before the question-specific cutoff; Profile questions use the complete timeline. Each framework uses its native memory construction and retrieval procedures to produce one free-form answer.

\textbf{Direct caption-memory baselines.}
The direct baselines convert the permitted video history into chronological clip-level records and answer without a retrieval stage. Gemini captions each complete clip, while the Qwen baselines instead caption 12 uniformly sampled frames per clip together with timestamped, speakerless ASR text~\citep{google2026gemini35flash,google2026gemini31pro,qwen3.5,qwen2026qwen3vl8b}. Their shared record schema captures the main event, visible actions and people, objects and places, dialogue or other audible content, and explicit temporal information when available.

\textbf{Memory and retrieval frameworks.}
Vgent constructs its memory graph at 1 fps and retrieves the five highest-ranked nodes for each question, after which its answerer receives the associated memory snippets and 12 frames from every selected clip~\citep{shen2025vgent}. M3-Agent uses the released checkpoint and official preprocessing~\citep{long2026m3agent}. Its Qwen2.5-Omni memorization model samples each input clip at 2 fps, while the auxiliary face-processing path samples frames at 5 fps for face detection and identity clustering~\citep{xu2025qwen25omni}. The reported M3-Agent ablations retain the same memory construction and control flow while disabling only face processing or face--voice equivalence linking.

\textbf{Transcripts and evaluation.}
Speaker-attribution labels are removed from the ASR transcripts, while names spoken in the dialogue are retained. The ASR-only controls retain these timestamped transcripts but remove frames, visual captions, face cues, and image-derived event descriptions. All model variants use the answerers reported in the main results table. The primary Gemini 3 Flash semantic judge receives only the question, reference answer, and system response, and runs at temperature $10^{-6}$~\citep{google2026gemini3flash}.

\section{Visual references used in construction}
\label{app:construction-examples}

\subsection{Main-theme construction artifacts}

\begin{figure*}[!t]
\begingroup
\captionsetup{type=table}
\centering
\small
\setlength{\tabcolsep}{4pt}
\renewcommand{\arraystretch}{1.0}
\begin{tabular}{@{}L{0.17\linewidth}L{0.30\linewidth}L{0.45\linewidth}@{}}
\toprule
Anchor & Visible cast and blocking & Scripted dialogue, audio, and camera \\
\midrule
A1: Kitchen &
Li Ming pours water at the counter, Wang Lin holds the mugs, and Zhang Hua watches from behind. &
Wang Lin: ``Li Ming, this glass V60 dripper makes the best holiday coffee.'' Soft bubbling water, glass clinks, and quiet morning ambience accompany a static observational shot. \\
\addlinespace
A2: Balcony &
Li Ming waters the succulents while Zhang Hua leans closer; Li Jian and Chen Tao watch nearby. &
Li Ming: ``Mom, look at how well this jade plant is doing this spring.'' Water, morning city ambience, and distant birdsong accompany a slow pan following the watering can. \\
\addlinespace
A5: Dining area &
Zhang Hua demonstrates how to tie a zongzi while Wang Lin observes; Li Jian and Li Ming stand behind them. &
Zhang Hua: ``Lin, keep the string tight so the rice doesn't leak out.'' Bamboo-leaf rustling and apartment ambience accompany a static, table-height camera. \\
\bottomrule
\end{tabular}
\caption{\textbf{Representative structured shot-plan excerpts.}
Anchors A1, A2, and A5 correspond to \texttt{clip\_001\_shot\_01}, \texttt{clip\_003\_shot\_01}, and \texttt{clip\_078\_shot\_01}; the rows condense their original cast, blocking, dialogue, audio, and camera fields.}
\label{tab:shot-plan-examples}
\endgroup
\vspace{6pt}
\centering
\footnotesize
\textbf{(a) Canonical references for the recurring cast}
\par\vspace{3pt}
\begin{minipage}[t]{0.154\linewidth}\centering
\includegraphics[width=\linewidth]{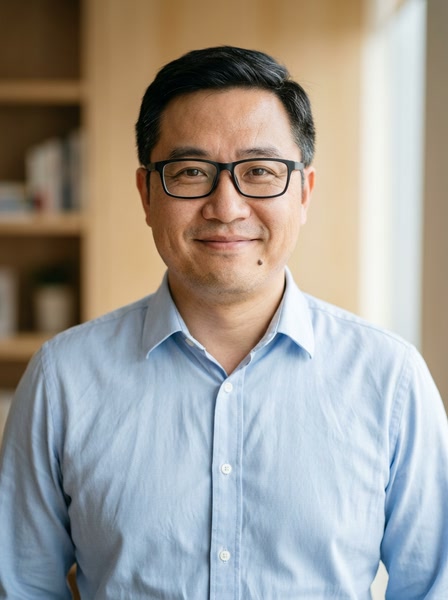}\\[-1pt]
\scriptsize Li Ming
\end{minipage}\hfill
\begin{minipage}[t]{0.154\linewidth}\centering
\includegraphics[width=\linewidth]{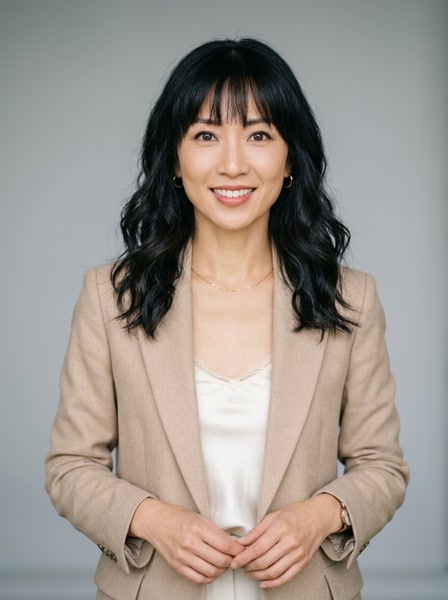}\\[-1pt]
\scriptsize Wang Lin
\end{minipage}\hfill
\begin{minipage}[t]{0.154\linewidth}\centering
\includegraphics[width=\linewidth]{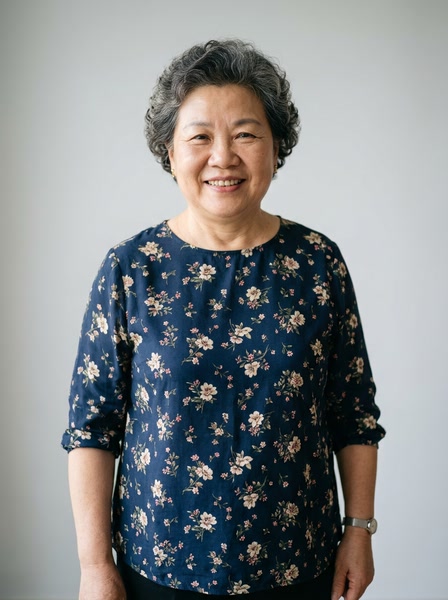}\\[-1pt]
\scriptsize Zhang Hua
\end{minipage}\hfill
\begin{minipage}[t]{0.154\linewidth}\centering
\includegraphics[width=\linewidth]{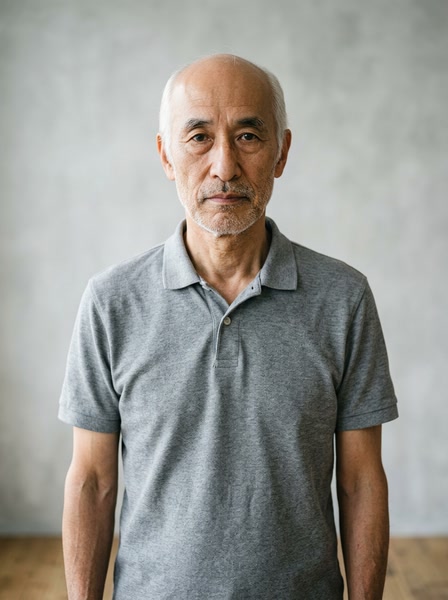}\\[-1pt]
\scriptsize Li Jian
\end{minipage}\hfill
\begin{minipage}[t]{0.154\linewidth}\centering
\includegraphics[width=\linewidth]{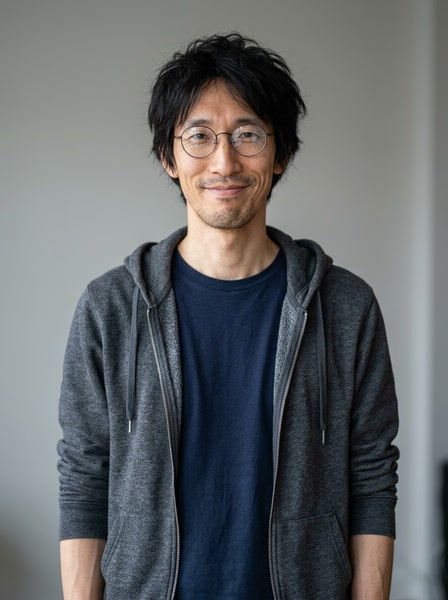}\\[-1pt]
\scriptsize Chen Tao
\end{minipage}\hfill
\begin{minipage}[t]{0.154\linewidth}\centering
\includegraphics[width=\linewidth]{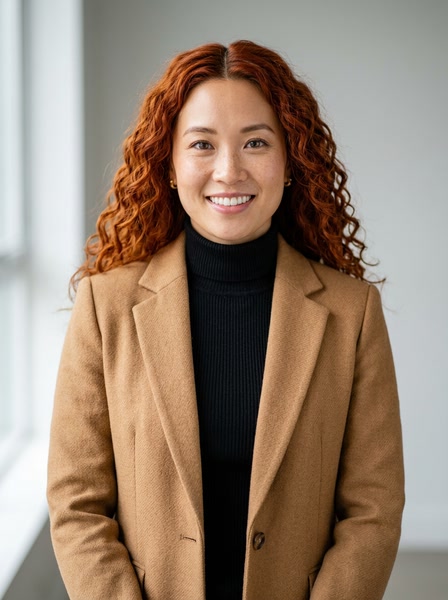}\\[-1pt]
\scriptsize Sarah Wu
\end{minipage}

\par\vspace{4pt}
\textbf{(b) Selected shot anchors across planned wardrobes and settings}
\par\vspace{3pt}
\begin{minipage}[t]{0.32\linewidth}\centering
\includegraphics[width=\linewidth]{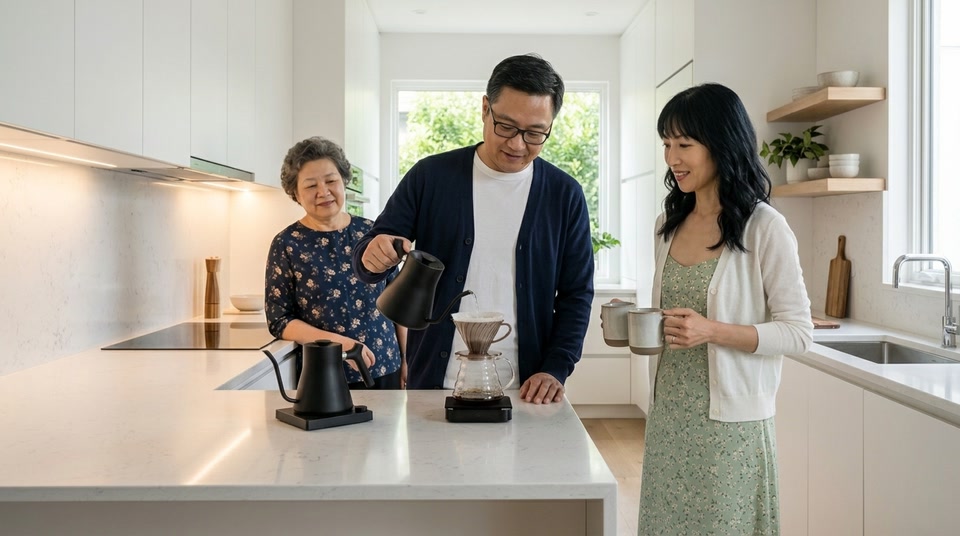}\\[-1pt]
\scriptsize A1: Holiday coffee
\end{minipage}\hfill
\begin{minipage}[t]{0.32\linewidth}\centering
\includegraphics[width=\linewidth]{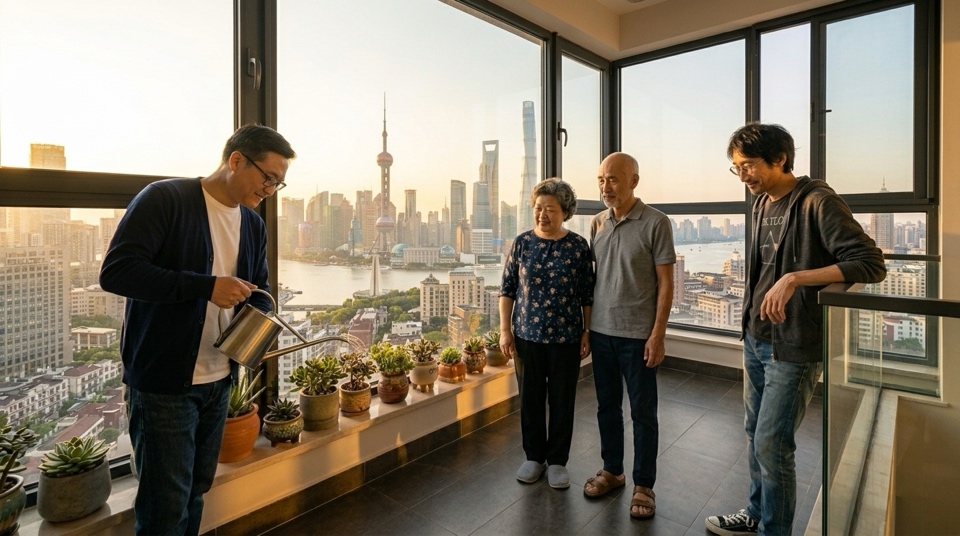}\\[-1pt]
\scriptsize A2: Balcony gardening
\end{minipage}\hfill
\begin{minipage}[t]{0.32\linewidth}\centering
\includegraphics[width=\linewidth]{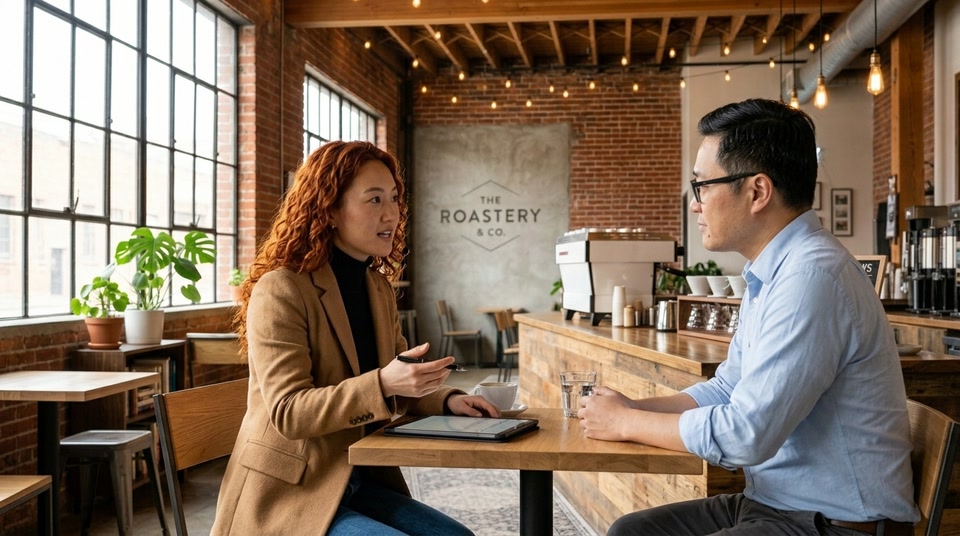}\\[-1pt]
\scriptsize A3: Caf\'e review
\end{minipage}

\par\vspace{2pt}
\begin{minipage}[t]{0.32\linewidth}\centering
\includegraphics[width=\linewidth]{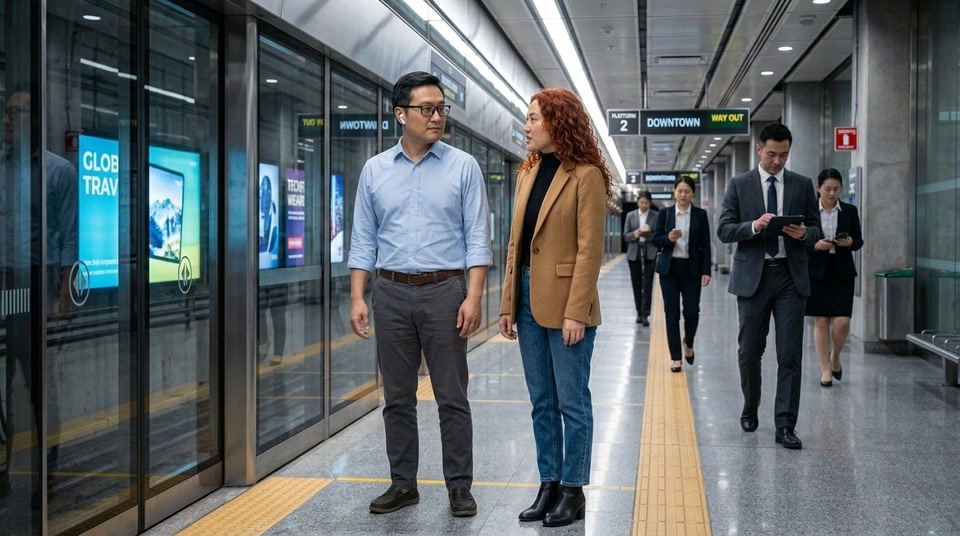}\\[-1pt]
\scriptsize A4: Transit commute
\end{minipage}\hfill
\begin{minipage}[t]{0.32\linewidth}\centering
\includegraphics[width=\linewidth]{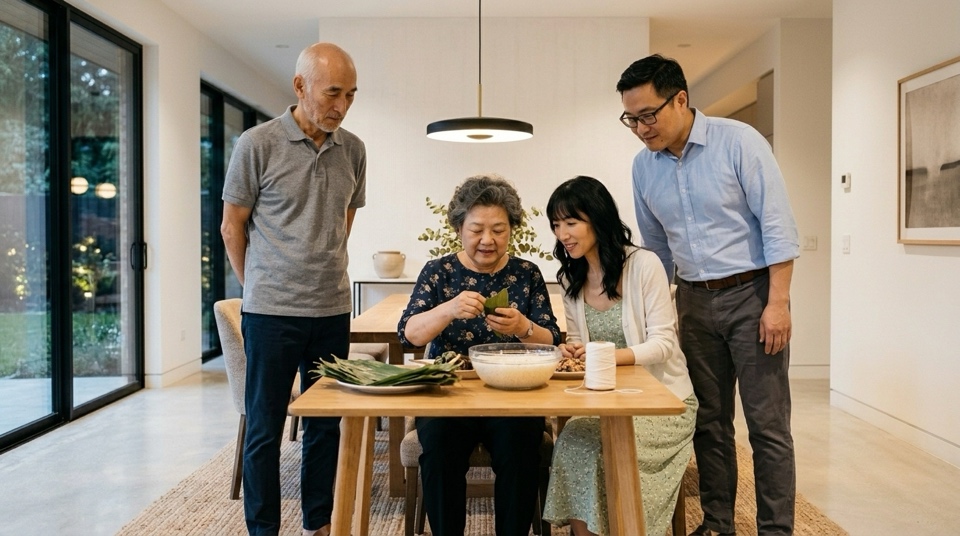}\\[-1pt]
\scriptsize A5: Family zongzi lesson
\end{minipage}\hfill
\begin{minipage}[t]{0.32\linewidth}\centering
\includegraphics[width=\linewidth]{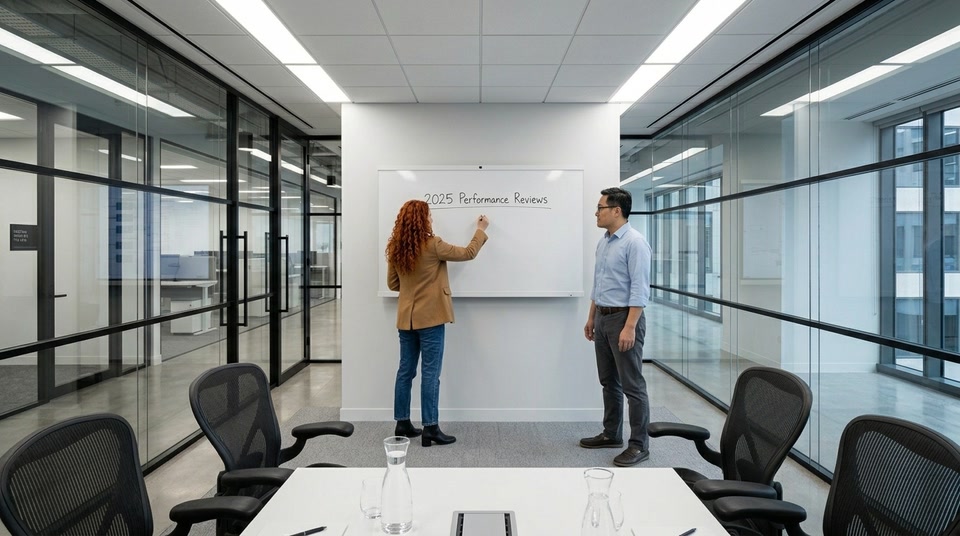}\\[-1pt]
\scriptsize A6: Workplace review
\end{minipage}
\caption{\textbf{Representative visual references used during construction.}
(a) Canonical images define the six recurring adults. (b) Selected anchors place those identities in planned wardrobes, activities, and settings before image-to-video rendering, including both paired and multi-person scenes.}
\label{fig:visual-reference-examples}
\vspace{-8pt}
\end{figure*}

Table~\ref{tab:shot-plan-examples} gives three records from the actual shot plans, including dialogue, audio, and camera instructions. Figure~\ref{fig:visual-reference-examples} then shows the corresponding visual references used before video rendering: canonical images establish the six recurring identities, while shot anchors combine the planned cast, wardrobe, setting, blocking, and composition.

\subsection{Second-theme construction example}
\label{app:second-theme-example}

\begin{figure*}[!t]
\begingroup
\captionsetup{type=table}
\centering
\small
\setlength{\tabcolsep}{4pt}
\renewcommand{\arraystretch}{1.0}
\begin{tabular}{@{}L{0.17\linewidth}L{0.36\linewidth}L{0.39\linewidth}@{}}
\toprule
Anchor & Visible cast and blocking & Scripted dialogue, audio, and camera \\
\midrule
S2: Stage rehearsal &
Thiago faces Andreia at the drum kit while Renata observes from the wing, with Clara and Felipe beside a costume rack. &
Thiago: ``Andreia, the tempo is way too fast for this monologue!'' A sharp drum-rim hit and theater echo accompany a static wide shot from the front of the stage. \\
\addlinespace
S4: Morning script review &
Renata reviews a script binder at the kitchen table while Carlos prepares the espresso machine at the counter behind her. &
Carlos: ``Ready for the first read-through today, Renata?'' Espresso hum, a clicking highlighter cap, and morning street sounds accompany a static medium shot with shallow depth of field. \\
\addlinespace
S6: Next-script caf\'e meeting &
Renata sits at the center of a small caf\'e table, with Carlos to her left and Thiago to her right, and signals to the off-camera barista. &
Renata: ``Mais um pingado, por favor, and maybe one for Carlos too.'' Soft caf\'e chatter and an espresso-machine hiss accompany a static medium shot. \\
\bottomrule
\end{tabular}
\caption{\textbf{Second-theme shot-plan excerpts.}
Anchors S2, S4, and S6 correspond to \texttt{clip\_14\_shot\_01}, \texttt{clip\_01\_shot\_01}, and \texttt{clip\_30\_shot\_01}; the rows condense their original cast, blocking, dialogue, audio, and camera fields.}
\label{tab:second-theme-shot-plan}
\endgroup
\vspace{6pt}
\centering
\footnotesize
\textbf{(a) Canonical references for the recurring cast}
\par\vspace{3pt}
\begin{minipage}[t]{0.154\linewidth}\centering
\includegraphics[width=\linewidth]{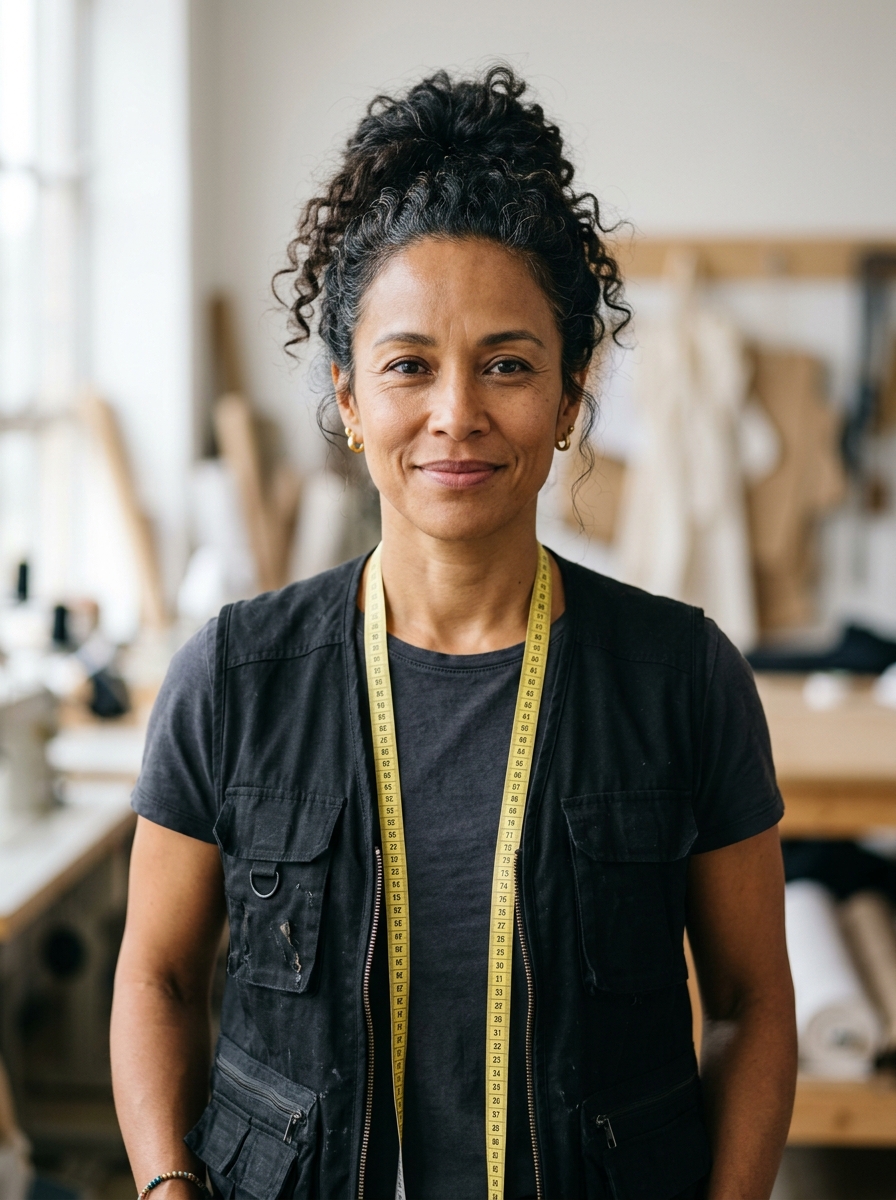}\\[-1pt]
\scriptsize Renata Silva
\end{minipage}\hfill
\begin{minipage}[t]{0.154\linewidth}\centering
\includegraphics[width=\linewidth]{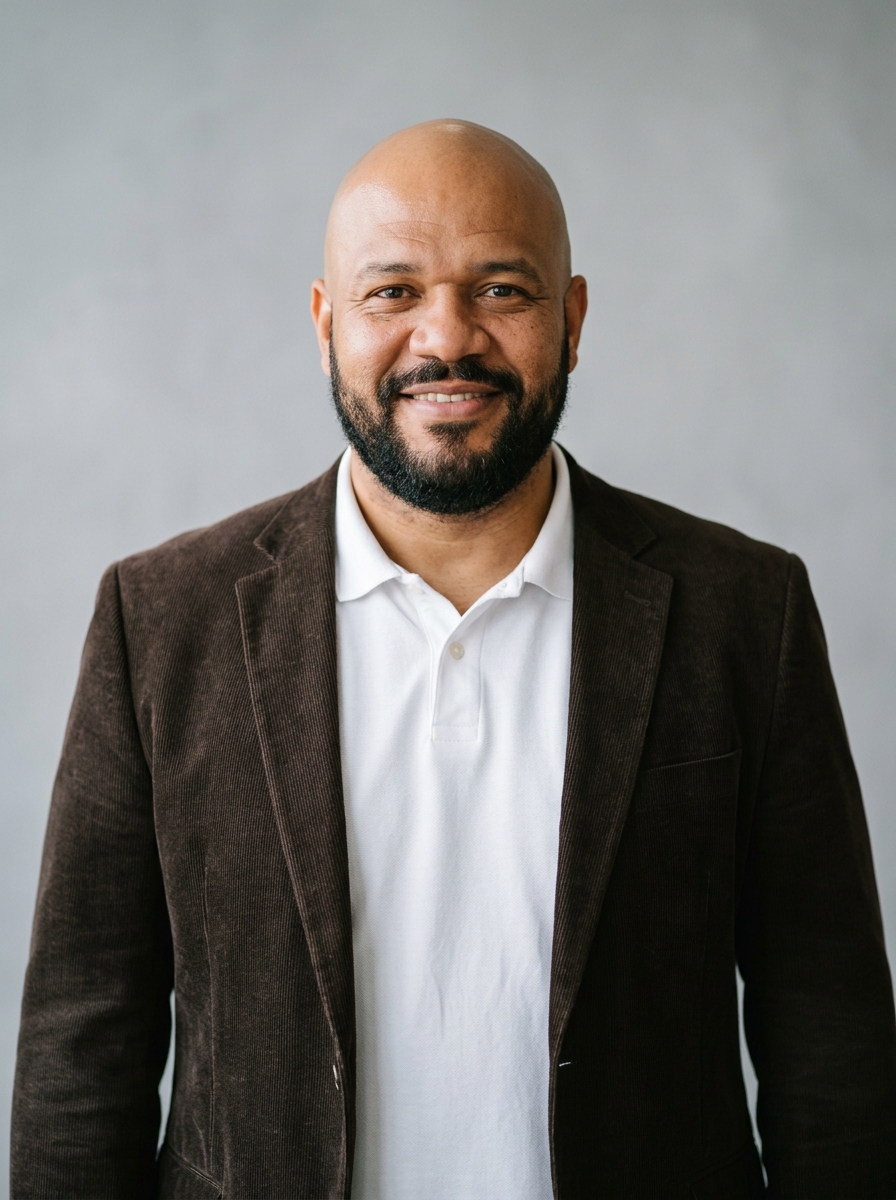}\\[-1pt]
\scriptsize Carlos Silva
\end{minipage}\hfill
\begin{minipage}[t]{0.154\linewidth}\centering
\includegraphics[width=\linewidth]{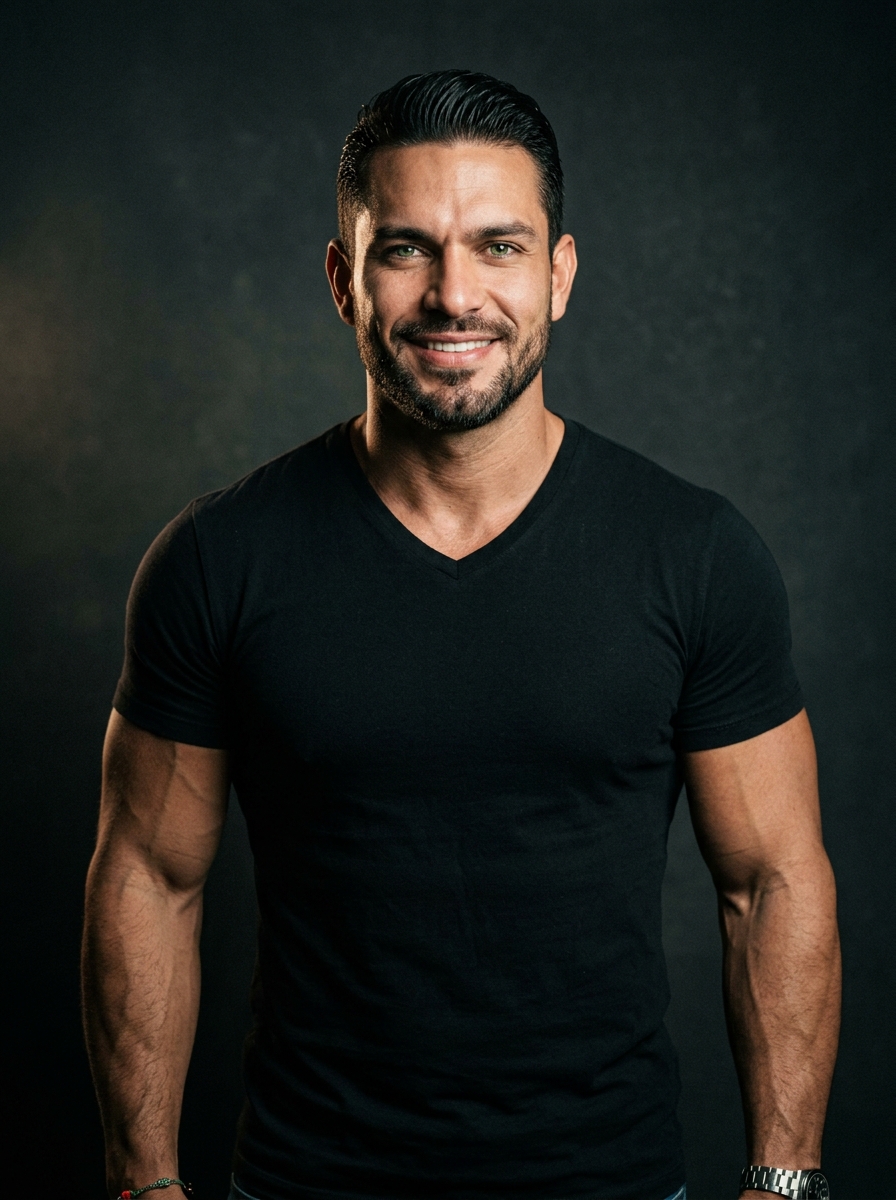}\\[-1pt]
\scriptsize Thiago Mendes
\end{minipage}\hfill
\begin{minipage}[t]{0.154\linewidth}\centering
\includegraphics[width=\linewidth]{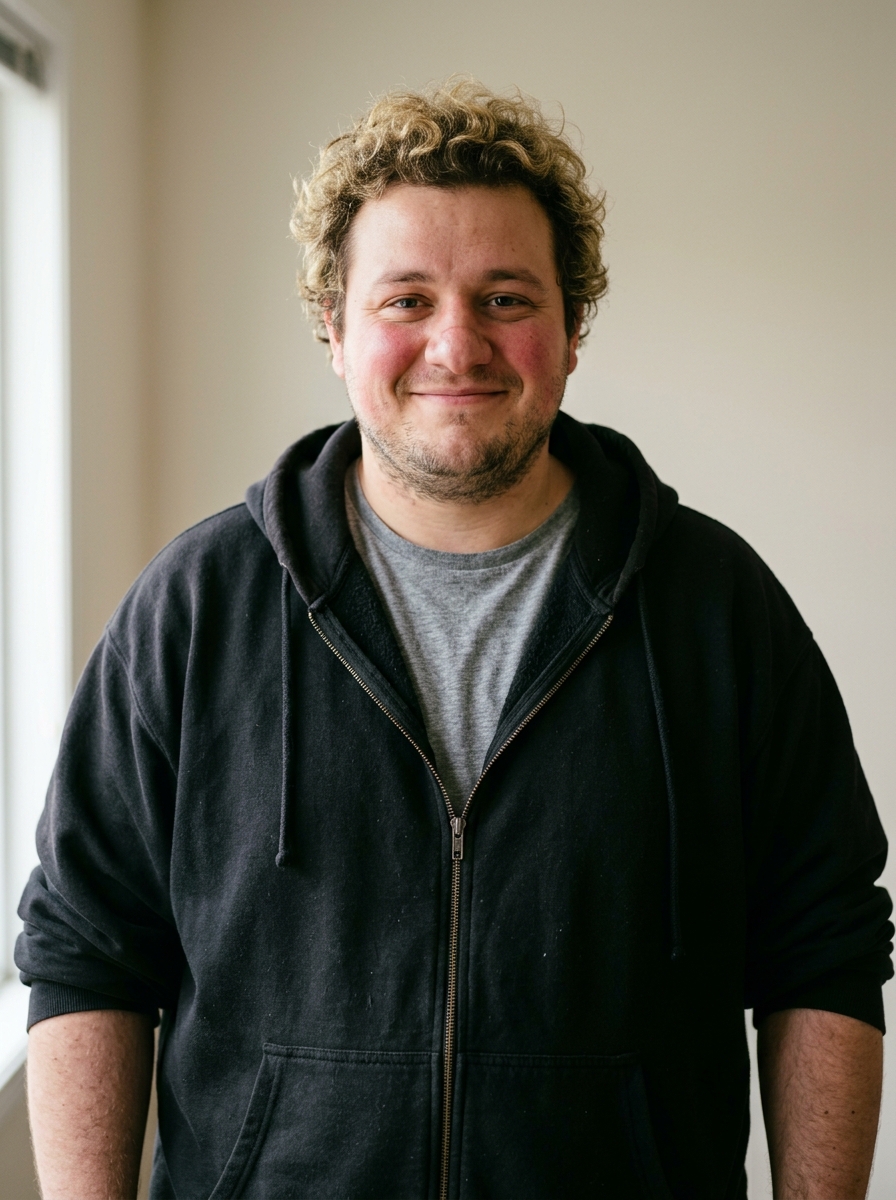}\\[-1pt]
\scriptsize Felipe Santos
\end{minipage}\hfill
\begin{minipage}[t]{0.154\linewidth}\centering
\includegraphics[width=\linewidth]{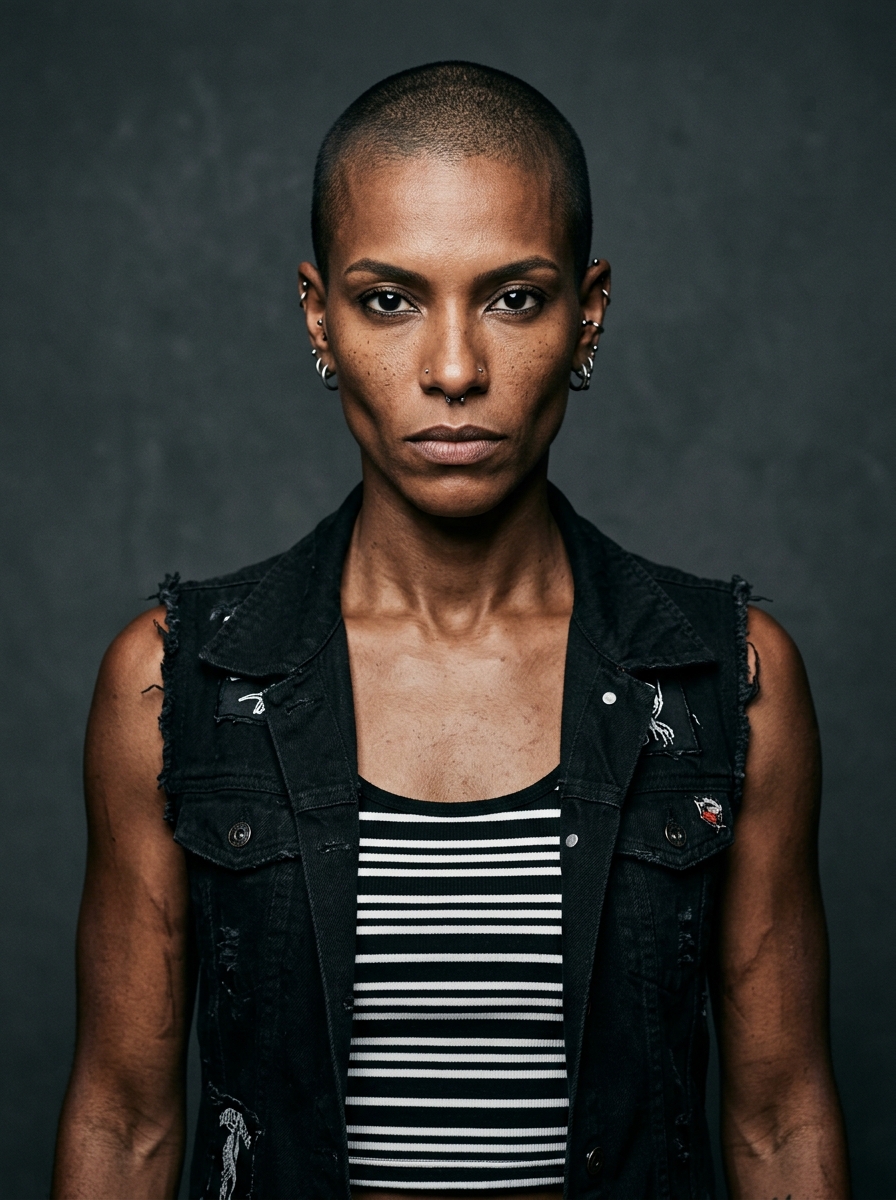}\\[-1pt]
\scriptsize Andreia Gomes
\end{minipage}\hfill
\begin{minipage}[t]{0.154\linewidth}\centering
\includegraphics[width=\linewidth]{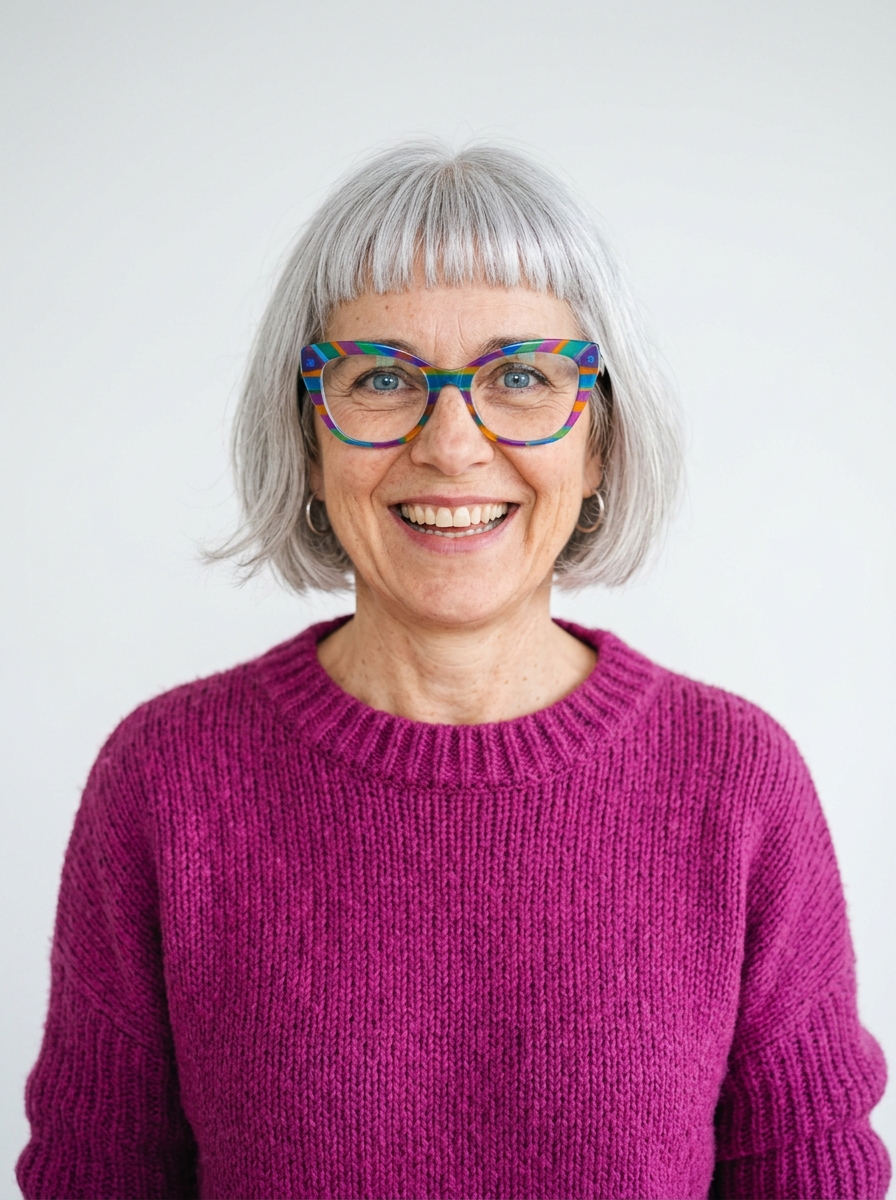}\\[-1pt]
\scriptsize Clara Oliveira
\end{minipage}

\par\vspace{4pt}
\textbf{(b) Selected shot anchors across planned wardrobes and settings}
\par\vspace{3pt}
\begin{minipage}[t]{0.32\linewidth}\centering
\includegraphics[width=\linewidth]{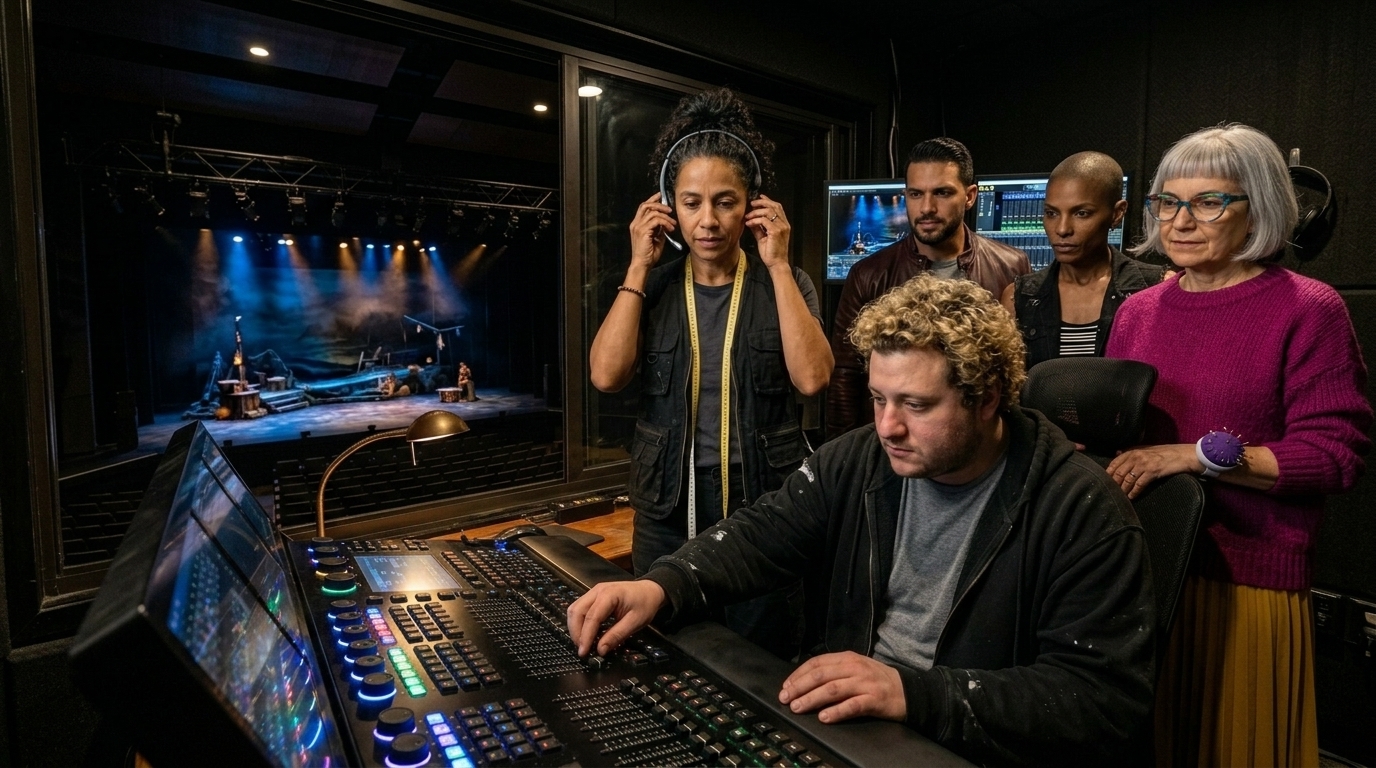}\\[-1pt]
\scriptsize S1: Control booth
\end{minipage}\hfill
\begin{minipage}[t]{0.32\linewidth}\centering
\includegraphics[width=\linewidth]{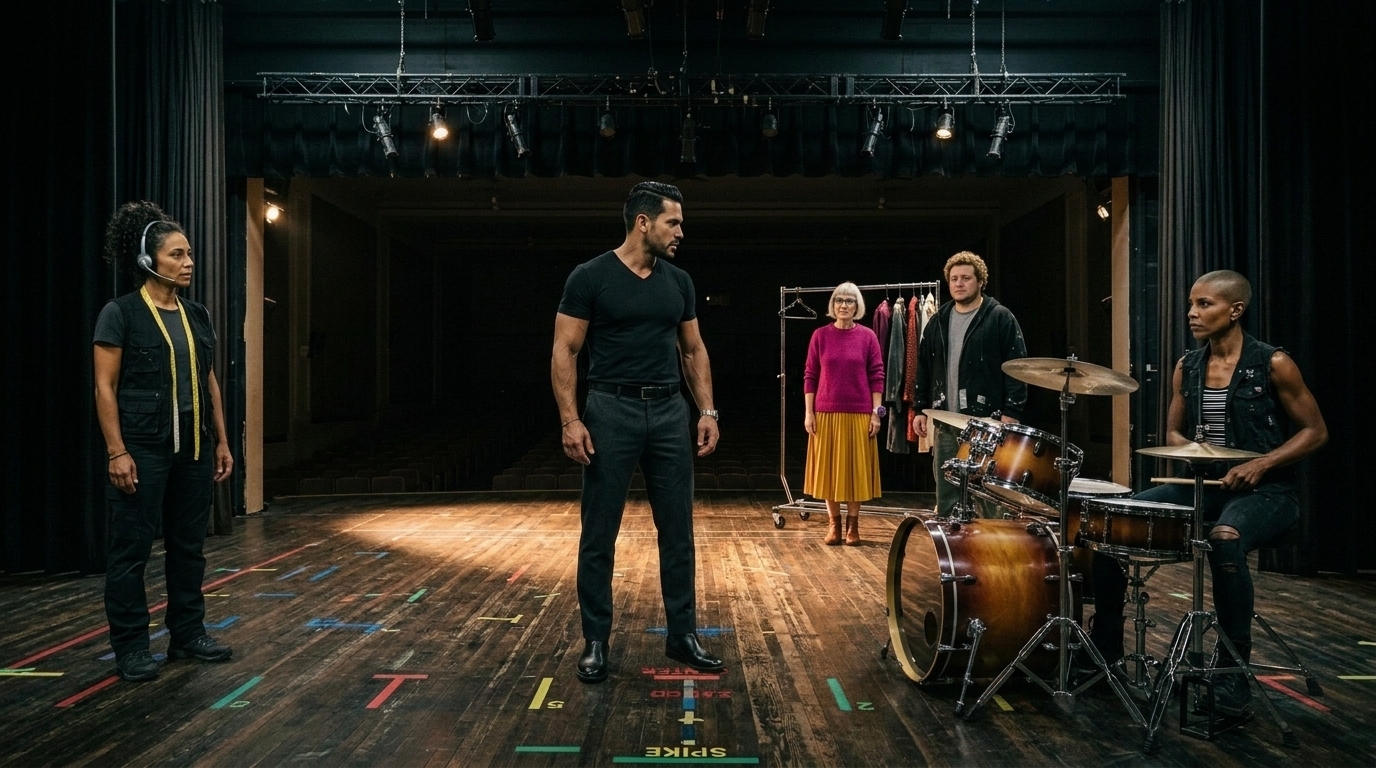}\\[-1pt]
\scriptsize S2: Stage rehearsal
\end{minipage}\hfill
\begin{minipage}[t]{0.32\linewidth}\centering
\includegraphics[width=\linewidth]{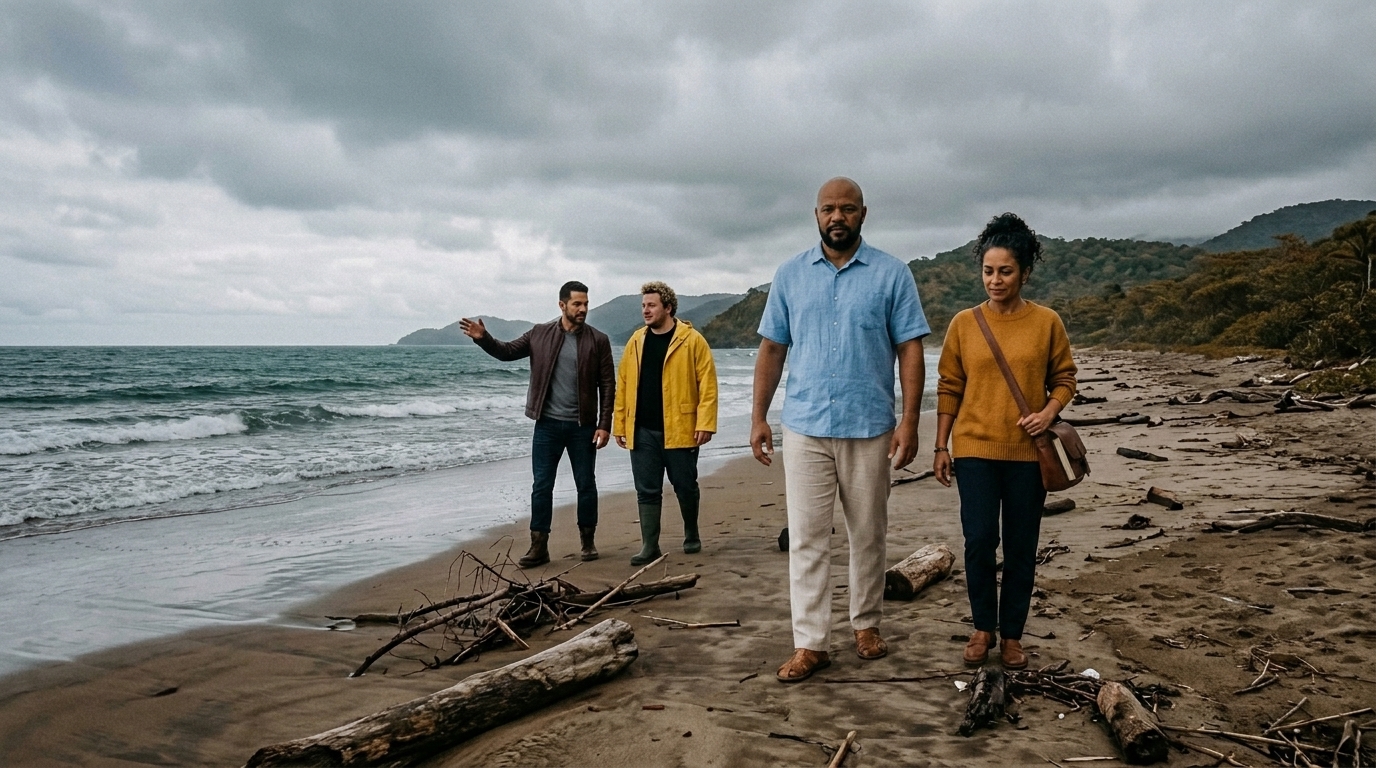}\\[-1pt]
\scriptsize S3: Coastal rest day
\end{minipage}

\par\vspace{2pt}
\begin{minipage}[t]{0.32\linewidth}\centering
\includegraphics[width=\linewidth]{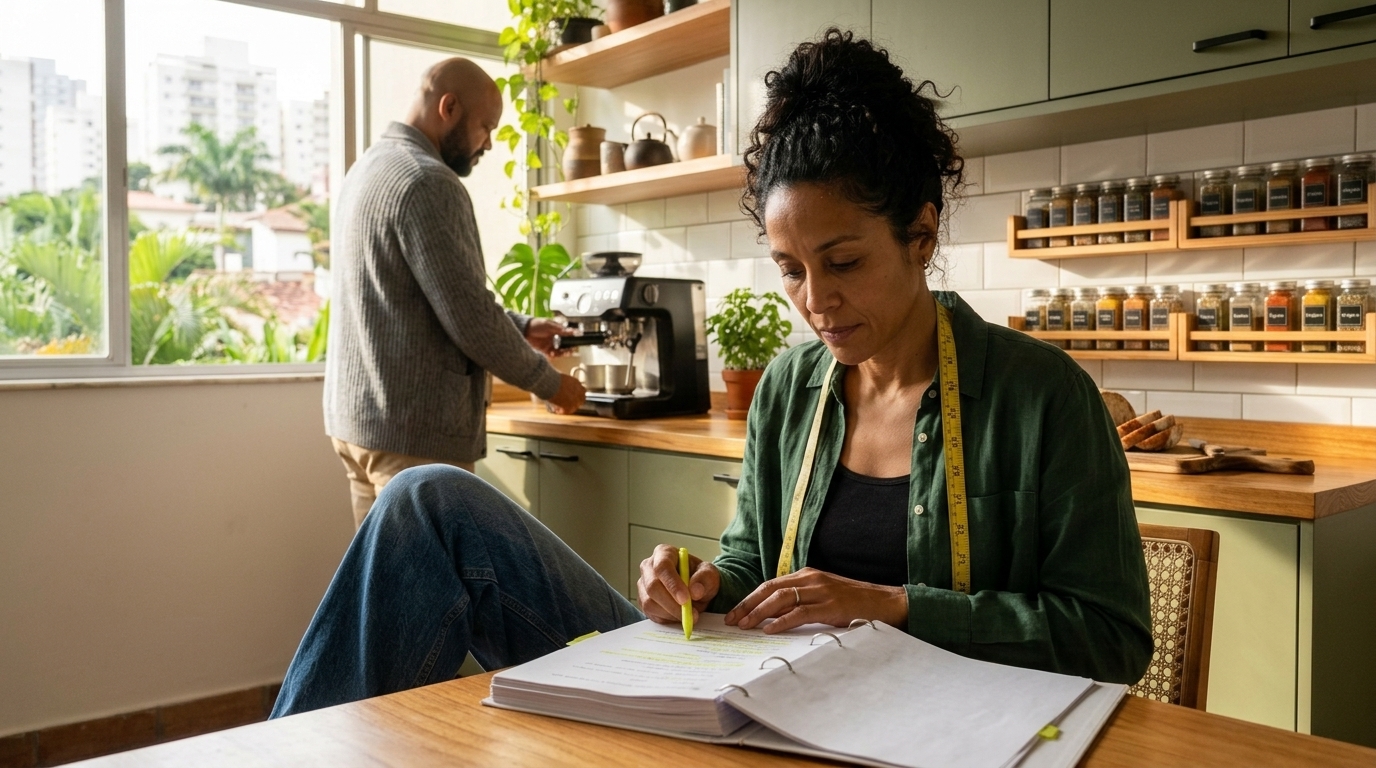}\\[-1pt]
\scriptsize S4: Morning script review
\end{minipage}\hfill
\begin{minipage}[t]{0.32\linewidth}\centering
\includegraphics[width=\linewidth]{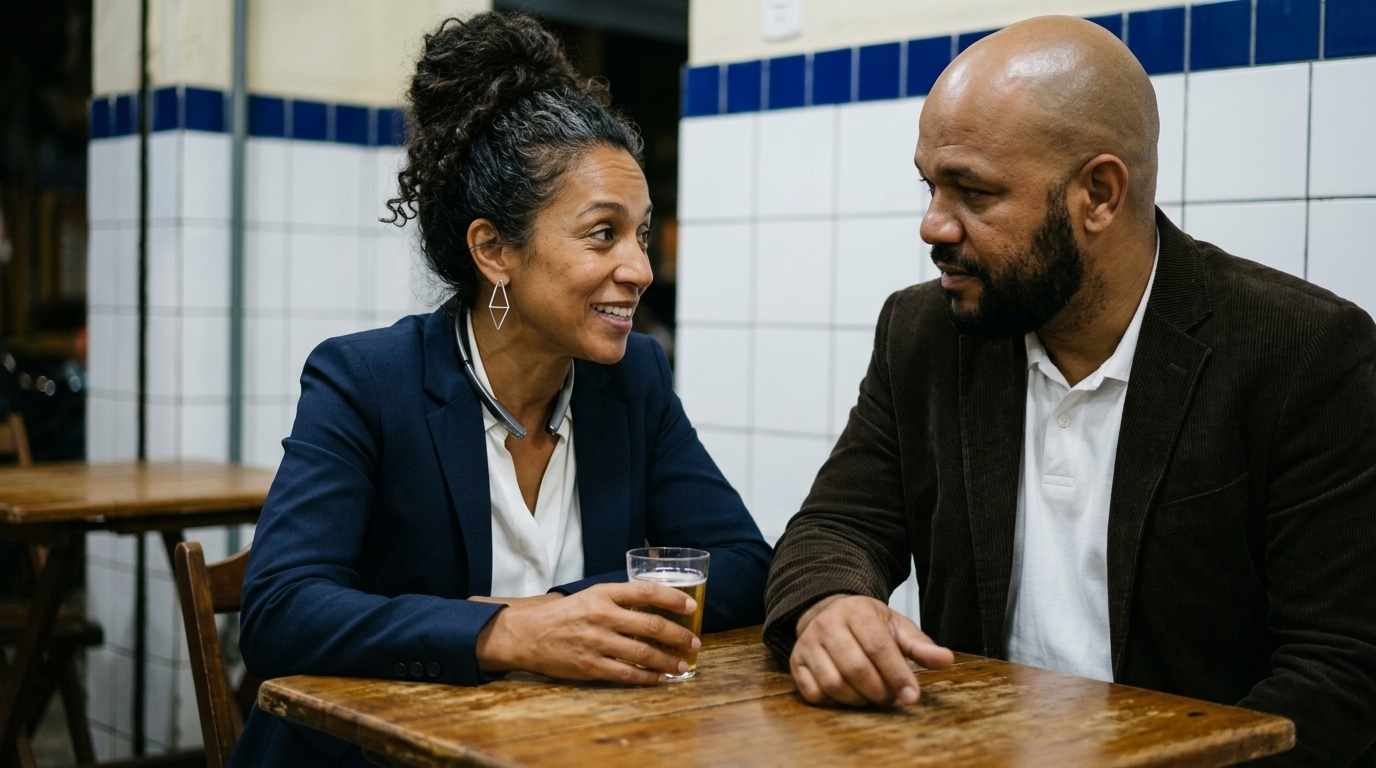}\\[-1pt]
\scriptsize S5: Opening-night toast
\end{minipage}\hfill
\begin{minipage}[t]{0.32\linewidth}\centering
\includegraphics[width=\linewidth]{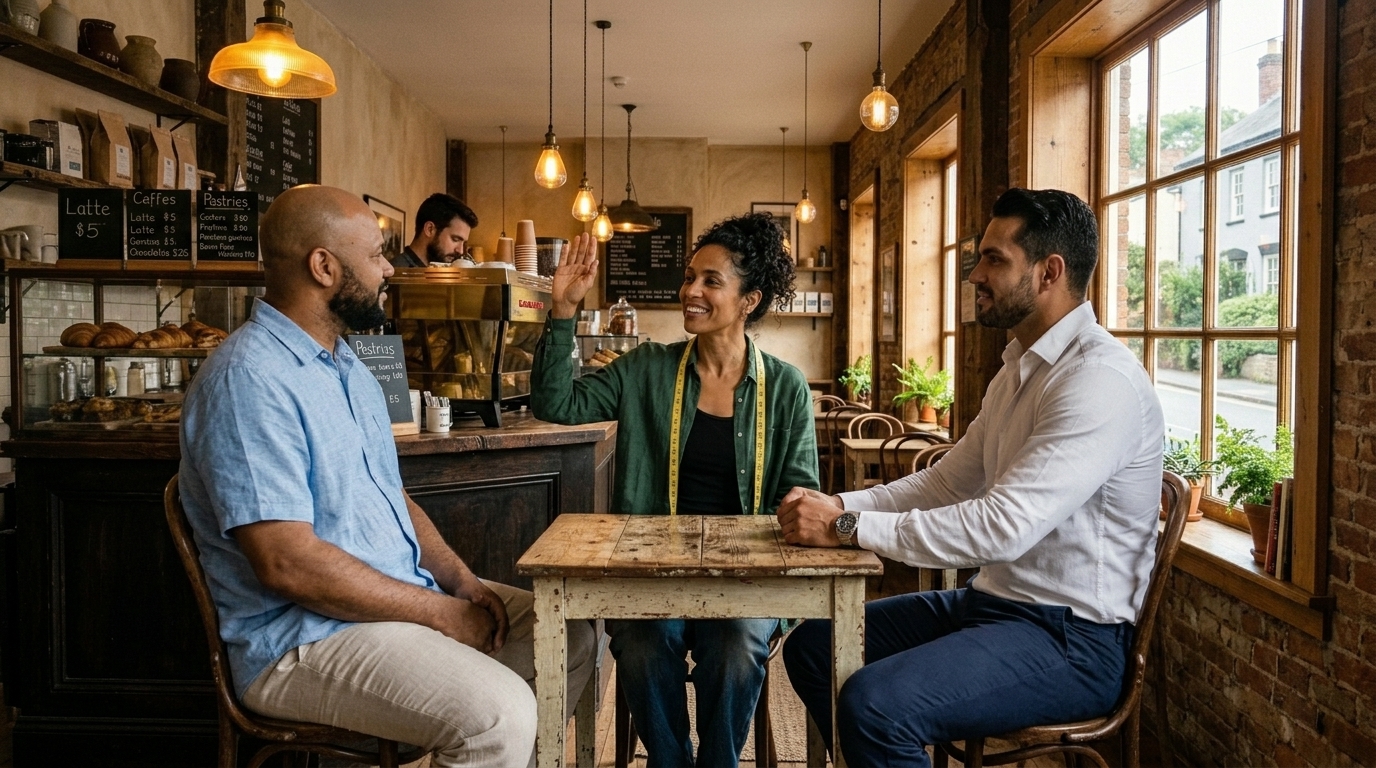}\\[-1pt]
\scriptsize S6: Next-script caf\'e meeting
\end{minipage}
\caption{\textbf{Second-theme construction example.}
(a) Canonical references for the six recurring adults in \textit{Backstage in S\~ao Paulo}. (b) Six pre-rendering shot anchors show these identities across theater work, rehearsal, outdoor rest, script preparation, and social settings. The same construction schema is reused for this one-month community-theater pilot.}
\label{fig:second-theme-examples}
\vspace{-8pt}
\end{figure*}

We also applied the construction process in Section~\ref{sec:generation-pipeline} to \textit{Backstage in S\~ao Paulo}, a one-month community-theater series with a new cast and setting. Table~\ref{tab:second-theme-shot-plan} gives three structured shot-plan records spanning stage rehearsal, a morning script review, and a caf\'e meeting. Figure~\ref{fig:second-theme-examples} shows six character references and six planned scenes, including these anchors. The pilot contains six recurring adults, 14 scene groups, 30 clips, and 84 planned shots. It demonstrates construction under a second theme and is not included in the benchmark QA or evaluation results.

\section{QA examples}
\label{app:qa-examples}

\benchmarkname{} questions are open-ended and are paired with lightweight evidence metadata.
Table~\ref{tab:app-qa-examples} shows selected examples from the three QA families, including questions that identify a queried participant through context or a family relation rather than always naming that person directly.

\begin{table*}[!tbp]
\centering
\small
\setlength{\tabcolsep}{3pt}
\begin{tabular}{@{}L{0.12\linewidth}L{0.51\linewidth}L{0.31\linewidth}@{}}
\toprule
Category & Question & Reference answer \\
\midrule
Recall &
Who is featured in the photo thumbnail that Li Ming discusses with his father while syncing his digital album? &
Wang Lin. \\
\addlinespace
Recall &
Who joins the Labor Day holiday walk while the new lens is being tested? &
Li Ming's parents, Zhang Hua and Li Jian, joined the Labor Day holiday walk. \\
\addlinespace
Recall &
What product does Sarah Wu use to try to fix her screeching office chair? &
WD-40. \\
\midrule
Retrieval &
Where did Li Ming temporarily hide the birthday gift that his wife later fastened around her neck while getting ready? &
In a zippered side pocket of his black leather briefcase. \\
\addlinespace
Retrieval &
When Li Ming uses a stylus and drawing tablet to brighten the pink petals in a photo of his wife, what action was she performing in that captured moment? &
Catching a falling cherry blossom petal. \\
\addlinespace
Retrieval &
Once the weather turns cold enough for his heavy winter coat, what tool does Li Ming use to clean it? &
A white sticky lint roller. \\
\midrule
Profile &
How does Li Ming's daily coffee brewing habit evolve as the seasons change throughout the year? &
He switches from hot pour-overs in the winter to iced coffee and cold brews during the summer heat. \\
\addlinespace
Profile &
How does Li Ming typically behave when sharing small treats, pastries, or street food with his wife? &
He instinctively prioritizes her enjoyment, consistently offering her the first bite or giving her the larger portion of the treat. \\
\addlinespace
Profile &
What role does Chen Tao frequently play during tense or demanding periods in the office? &
He breaks the tension and boosts team morale using humor, playful drawings, and by sharing premium snacks. \\
\bottomrule
\end{tabular}

\caption{\textbf{Selected open-ended QA examples.}
Recall questions target a local memory, Retrieval questions connect evidence across multiple clips, and Profile questions aggregate repeated person-level evidence over time.}
\label{tab:app-qa-examples}
\vspace{-10pt}
\end{table*}

\section{QA validation and curation protocol}
\label{app:qa-quality-control}

\textbf{Automated validation.} We checked that every candidate contained a complete question, category, target identity, reference answer, and evidence record, and that all referenced clips belonged to the released timeline. Temporal and category checks required Recall questions to be grounded in a local event, Retrieval questions to connect at least two temporally separated clips within the permitted cutoff, and Profile questions to summarize a recurring role, habit, preference, or relationship pattern over the full timeline. We also removed malformed, semantically redundant, category-inconsistent, and answer-leaking candidates.

\textbf{Human curation.} Two authors reviewed every retained question against the permitted audio-visual memory. Before consulting the reference answer or evidence metadata, they assessed whether the question was answerable from the available video. They then checked the target identity, answer, supporting evidence, category, and cutoff, rejecting items with insufficient evidence, ambiguous targets, unnatural wording, or multiple incompatible answers supported by the available evidence. Disagreements were resolved by rechecking the relevant clips. Only questions that passed every check entered the final set of 1,217; this curation was separate from the post-hoc answerability exercise in Appendix~\ref{app:human-reference}.

\section{Human answerability reference}
\label{app:human-reference}

\looseness=2
To estimate benchmark answerability, two independent human evaluators who were not involved in benchmark construction answered all 1,217 questions using only the audio-visual memory permitted for each item. Recall and Retrieval questions exposed only clips up to the specified cutoff, whereas Profile questions used the full timeline. Both evaluators answered in an open-ended format without a time limit and had no access to the reference answers, evidence annotations, metadata, or model outputs during evaluation. The authors cross-checked both human answer sets against the reference answers using the same correctness criterion applied to system outputs. The arithmetic mean of the two evaluator-level accuracies was 93.8\% overall, including 96.8\% on Recall, 94.2\% on Retrieval, and 89.6\% on Profile.

\section{Judge reliability protocol}
\label{app:judge-reliability}

The judge-reliability audit uses two complete answer runs: M3-Agent (Original) with Qwen2.5-Omni-7B (SFT) and the Gemini 3.1 Pro direct caption-memory baseline. Each run contains one response for every benchmark question, for 1,217 responses per run. Gemini 3 Flash and GPT-5.4-mini independently assign a binary correctness label to every response from the question, reference answer, and recorded system answer\vadjust{\vspace{0pt plus 1pt}} using the same semantic-equivalence prompt~\citep{google2026gemini3flash,openai2026gpt54mini}. After completing the answerability exercise,\vadjust{\vspace{0pt plus 1pt}} the same two independent human evaluators also verify every response in both runs\vadjust{\vspace{0pt plus 1pt}} using the question, reference answer, and system answer. Judge--human agreement is computed separately against each evaluator's labels and then averaged. For M3-Agent, mean agreement is 96.6\% for Gemini 3 Flash and 95.7\% for GPT-5.4-mini; for Gemini direct, the corresponding values are 96.1\% and 95.3\%.

\vspace{-6pt}
\section{Evaluation prompt templates}
\label{app:prompt-templates}
\vspace{-4pt}

We provide three prompt templates that are central to reproducing the evaluation protocol, covering caption-memory construction, open-ended answering, and semantic judging, while system-specific implementations use only the minimal wrappers required by each framework interface.

\Needspace{6\baselineskip}
\subsection{Caption generation}
\vspace{-4pt}

\begin{promptbox}[Qwen frame-and-ASR caption prompt]
System:
You are an expert video captioner for long-term personal memory QA.

User:
Caption this personal album video clip for later open-ended QA.

Clip id: {clip_id}

Transcript/ASR notes, without guaranteed speaker labels:
{transcript}

Return exactly one JSON object with these fields:
- clip_id: the provided clip id
- caption: 3-6 dense sentences summarizing visible actions, people, objects, setting, and any audible speech
- visual_details: short list of concrete visual evidence
- dialogue_or_speech: short list of exact or near-exact spoken facts if audible
- audio_events: short list of relevant non-speech audio, or []
- people: short list of people names/roles if identifiable, or []
- objects_places: short list of important objects and places
- time_or_event: date/time/event if stated or visually clear, otherwise ""

Rules:
- Use only what is visible in the sampled frames or supported by the transcript/ASR notes.
- Preserve specific names, objects, places, colors, dates, and dialogue facts.
- If speech is unclear, say it is unclear instead of inventing.
- Do not include markdown.
- Return valid JSON only.
\end{promptbox}

\Needspace{6\baselineskip}
\subsection{Open-ended answering}
\vspace{-4pt}

\begin{promptbox}[Open-ended answering prompt]
System:
You are a memory QA assistant. Use the provided evidence to answer.
Give the most likely concise answer. Do not answer 'Unknown'. Respond with only the answer.
If there is insufficient information, you can make reasonable guesses.
Do not output reasoning or analysis. /no_think

User:
Question: {question}

Evidence:
{evidence}

Provide the most likely concise answer using the evidence. If there is insufficient information, you can make reasonable guesses. /no_think
\end{promptbox}

\Needspace{6\baselineskip}
\subsection{Open-ended semantic judge}
\vspace{-4pt}

\begin{promptbox}[Semantic-judge prompt]
You are provided with a question, a ground truth answer, and an answer from an agent model. Your task is to determine whether the ground truth answer can be logically inferred from the agent's answer, in the context of the question.

Do not directly compare the surface forms of the agent answer and the ground truth answer. Instead, assess whether the meaning expressed by the agent answer supports or implies the ground truth answer. If the ground truth can be reasonably derived from the agent answer, return "Yes". If it cannot, return "No".

Important notes:
- Do not require exact wording or matching structure.
- Semantic inference is sufficient, as long as the agent answer entails or implies the meaning of the ground truth answer, given the question.
- Only return "Yes" or "No", with no additional explanation or formatting.

Input fields:
- question: the question asked
- ground_truth_answer: the correct answer
- agent_answer: the model's answer to be evaluated

Now evaluate the following input:
Input:
- question: {question}
- ground_truth_answer: {ground_truth_answer}
- agent_answer: {agent_answer}
Output ('Yes' or 'No'):
\end{promptbox}

\begin{figure}[!t]
\centering
\begin{minipage}[t]{\linewidth}
\centering
\textbf{(a) Detected face observations}\par\vspace{4pt}
\begin{minipage}[t]{0.47\linewidth}\centering
\includegraphics[height=2.25cm]{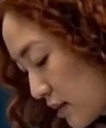}\\[-1pt]
\scriptsize Face observation A
\end{minipage}\hfill
\begin{minipage}[t]{0.47\linewidth}\centering
\includegraphics[height=2.25cm]{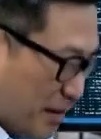}\\[-1pt]
\scriptsize Face observation B
\end{minipage}
\par\vspace{4pt}
\scriptsize The crops are taken directly from M3-Agent's face-preprocessing output for \texttt{clip\_276}.
\end{minipage}
\par\vspace{4pt}
\begin{minipage}[t]{\linewidth}
\textbf{(b) Voice linking and stored memory}\par\vspace{4pt}
\begin{tcolorbox}[icmbox,unbreakable,title={Voice features and generated equivalence}]
{\ttfamily\scriptsize
<voice\_3053>: ``Sarah, this kernel level leak is finally patched.''\\
<voice\_3886>: ``Great work. Does that clear the blocker?''\\
<voice\_945>: ``Yes, I'm marking it as resolved right now.''\\[2pt]
Equivalence: <voice\_3886>, <voice\_3053>\\
Mapped identifier: <character\_0> for all three voice nodes\par
}
\end{tcolorbox}
\vspace{3pt}
\begin{tcolorbox}[icmbox,unbreakable,title={Downstream clip-memory excerpt}]
{\ttfamily\scriptsize
<character\_0> and <character\_0> are working at a desk.\\
<character\_0> tells <character\_0>, ``Sarah, this kernel level leak is finally patched.''\\
<character\_0>'s name is Sarah.\par
}
\end{tcolorbox}
\end{minipage}
\caption{\textbf{Observed M3-Agent identity collision.}
Two distinct face observations and three voice segments come from the same evaluated clip. The saved equivalence and character mappings merge voice observations from different speakers, after which the downstream memory uses the same character identifier for both visible participants.}
\label{fig:m3-identity-collision}
\vspace{-8pt}
\end{figure}
\section{M3-Agent identity-linking diagnostic}
\label{app:m3-link-diagnostic}

M3-Agent represents people through face and voice features, predicts equivalence relations between these observations, and merges the linked features into persistent character identifiers~\citep{long2026m3agent}. In one evaluated run, the agent queries ``What is the character id of Li Ming.'' Most returned identity records associate Li Ming with \texttt{<character\_0>}, while one uses \texttt{<character\_17>}.

Figure~\ref{fig:m3-identity-collision} shows a second identity-linking error in public \texttt{clip\_276}. Face preprocessing detects two visually distinct principal participants, while the voice path extracts three spoken segments. In the saved graph, \texttt{<voice\_3053>}, \texttt{<voice\_3886>}, and \texttt{<voice\_945>} all map to \texttt{<character\_0>}; the generated equivalence record explicitly merges \texttt{<voice\_3886>} with \texttt{<voice\_3053>}. After character substitution, the stored clip memory consequently represents both visible participants as \texttt{<character\_0>} and assigns Sarah's name to that identifier.

\section{Detailed Failure Cases}
\label{app:failure-analysis}

This section presents three Profile failures, tracing each from its supporting evidence to the recorded system output.

\subsection{Relationship-specific profile miss}

\textbf{Question.} ``What does Li Ming usually try to capture when he photographs Wang Lin?''

\textbf{Evidence pattern.} Three retained episodes show Li Ming photographing Wang Lin's spontaneous reactions and interactions with her surroundings. The answer depends on connecting the recurring subject of these photographs to their relationship rather than merely recognizing common visual themes in the scenes.

\textbf{Reference answer.} Li Ming prioritizes candid, unposed moments of Wang Lin's natural joy and interactions with her environment.

\textbf{M3-Agent execution trace.} Starting with no retrieved knowledge, M3-Agent queries its memory for Li Ming and uses the dominant result, \texttt{<character\_0>}, to search for ``What type of photographic subject does \texttt{<character\_0>} prioritize when documenting shared experiences?'' The search returns two memories: one emphasizes a macro lens, city views, and leaf detail, while the other shows park and cat photography. From this context, the agent answers, ``Li Ming consistently prioritizes candid street photography over posed portraits when documenting shared experiences.'' The semantic judge marks this answer incorrect.

\textbf{Diagnosis.} The retrieval finds photography-related memories about Li Ming but misses that Wang Lin is the recurring subject. As a result, the answer describes a generic photographic style rather than Li Ming's repeated preference for candid photographs of Wang Lin.

\subsection{Retrieval distractor}

\textbf{Question.} ``What item do Li Ming and Wang Lin like collecting during their trips?''

\textbf{Evidence pattern.} Three temporally separated episodes show the couple examining postcards during a trip, purchasing additional cards, and organizing the collected cards afterward. The answer requires identifying the object that recurs across all three episodes.

\textbf{Reference answer.} Postcards.

\begin{table}[!htbp]
\centering
\small
\setlength{\tabcolsep}{3pt}
\begin{tabular}{@{}r l r L{0.50\linewidth}@{}}
\toprule
Rank & \shortstack[l]{Public\\clip} & Score & Retrieved memory cue \\
\midrule
1 & \texttt{clip\_181} & 0.5614 & Memory cards organized in a hard case for travel \\
2 & \texttt{clip\_368} & 0.5508 & A digital frame displaying a photograph from a trip \\
3 & \texttt{clip\_375} & 0.5498 & Seasonal clothing being packed away for winter \\
4 & \texttt{clip\_256} & 0.5421 & A passport pouch prepared for a Europe trip \\
5 & \texttt{clip\_365} & 0.5417 & Train tickets and European receipts stored in a folder \\
\bottomrule
\end{tabular}

\caption{\textbf{Vgent + Qwen3.5-9B retrieval.} The run returns the following five memories in descending similarity order. Clip identifiers follow the public dataset release.}
\vspace{-8pt}
\end{table}

\begin{table}[!htbp]
\centering
\small
\setlength{\tabcolsep}{3pt}
\begin{tabular}{@{}L{0.45\linewidth}L{0.34\linewidth}L{0.14\linewidth}@{}}
\toprule
System & Recorded answer & Correct \\
\midrule
Gemini 3.1 Pro direct & Postcards. & Yes \\
Vgent + Qwen3.5-9B & Travel receipts or tickets. & No \\
Vgent + Qwen3-VL-8B-Instruct & Memory cards. & No \\
HippoRAG2 + Qwen3.5-9B & Sepia-toned postcards. & Yes \\
HippoRAG2 + Qwen3-VL-8B-Instruct & Postcards. & Yes \\
\bottomrule
\end{tabular}

\caption{\textbf{Recorded system answers.}}
\vspace{-8pt}
\end{table}

\textbf{Diagnosis.} None of the five retrieved memories is one of the annotated postcard episodes (public clips \texttt{clip\_197}, \texttt{clip\_357}, and \texttt{clip\_389}). The returned set instead contains several travel and record-keeping distractors, and the final memory explicitly mentions tickets and receipts. Vgent consequently repeats that distractor in its answer.

\subsection{Incomplete cross-clip evidence integration}

\textbf{Question.} ``What gardening tools and accessories does Li Ming repeatedly use to maintain the hydration and health of his balcony succulents and ferns?''

\textbf{Evidence pattern.} The selected memories include pruning, watering, and hydration during travel. Public \texttt{clip\_288} mentions silver scissors used for a fern, and \texttt{clip\_410} describes a copper watering can for succulents. In \texttt{clip\_323}, the retrieved text explicitly states, ``These glass bulbs should keep the ferns hydrated while we're away.''

\textbf{Reference answer.} He uses long-spouted watering cans, spray bottles for misting leaves, pruning shears for dead stems, and glass auto-watering bulbs for extended absences.

\begin{table}[!htbp]
\centering
\small
\setlength{\tabcolsep}{3pt}
\begin{tabular}{@{}L{0.36\linewidth}L{0.43\linewidth}L{0.14\linewidth}@{}}
\toprule
System & Recorded answer & Correct \\
\midrule
Vgent + Qwen3-VL-8B-Instruct & Li Ming repeatedly uses a glass watering can and silver shears to maintain the hydration and health of his balcony succulents and ferns. & No \\
\bottomrule
\end{tabular}

\caption{\textbf{Recorded system answer.}}
\vspace{-8pt}
\end{table}

\Needspace{8\baselineskip}
\textbf{Diagnosis.} The answer retains watering and pruning tools but omits the auto-watering bulbs explicitly described in the retrieved evidence. This illustrates incomplete integration of available details across clips into a person-level answer.

\end{document}